\documentclass{article}

 \usepackage[preprint]{neurips_2026}

\usepackage[utf8]{inputenc} 
\usepackage[T1]{fontenc}    
\usepackage{hyperref}

\usepackage{url}            
\usepackage{booktabs}       
\usepackage{amsfonts}       
\usepackage{nicefrac}       
\usepackage{microtype}      
\usepackage{xcolor}         

\usepackage{graphicx}
\usepackage{array}
\usepackage[export]{adjustbox}
\usepackage{subcaption}

\usepackage{booktabs}
\usepackage{makecell}
\usepackage{wrapfig}
\usepackage{amssymb}
\usepackage{pifont}
\usepackage{tabularx}
\usepackage{csquotes}
\usepackage[table]{xcolor}
\usepackage[most]{tcolorbox}
\usepackage{longtable}
\usepackage{cleveref}
\usepackage{fontawesome}

\definecolor{OIblue}{HTML}{0072B2}
\definecolor{OIorange}{HTML}{E69F00}
\definecolor{OIgray}{gray}{0.25}

\usepackage[para,online,flushleft]{threeparttable}

\newcommand{\cmark}{\textcolor{green!60!black}{\ding{51}}}
\newcommand{\xmark}{\textcolor{red!60!black}{\ding{55}}}  

\title{BuilderBench: The Building Blocks of \\ Intelligent Agents}

\author{Raj Ghugare\textsuperscript{1} \quad Roger Creus Castanyer\textsuperscript{2,3} \quad Catherine Ji\textsuperscript{1} \quad Kathryn Wantlin\textsuperscript{1} \\
[0.2cm]
\textbf{Jin Schofield\textsuperscript{1} \quad Karthik Narasimhan\textsuperscript{*\ 1} \quad Benjamin Eysenbach\textsuperscript{*\ 1}}
\\[0.2cm]
\textsuperscript{1}{Princeton University} \quad \textsuperscript{2}{Mila – Quebec AI Institute} \quad \textsuperscript{3}{Universit\'e de Montr\'eal}
}

\begin{document}

\maketitle

\begin{abstract}
Today's AI models learn primarily through mimicry and refining, so it is not surprising that they struggle to solve problems beyond the limits set by existing data. To solve novel problems, agents should acquire skills by exploring and learning through experience. Finding a scalable learning mechanism for developing agents that learn through interaction remains a major open problem. In this work, we introduce BuilderBench, a benchmark to accelerate research into agent training that centers open-ended exploration. BuilderBench requires agents to learn how to build any structure using blocks. BuilderBench is equipped with $(1)$ a simulator of a robot interacting with various physical blocks, and $(2)$ a task-suite with over 50 diverse target structures that are carefully curated to test an understanding of physics, mathematics, and long-horizon planning. Agents are provided with a target structure at the start, and can interact with the environment for multiple episodes to experiment and learn various skills for building the structure. Solving these tasks requires \emph{embodied reasoning} in a way that is not reflected in words but rather in actions, experimenting with different strategies and piecing them together. 
Our experiments with multiple state-of-the-art frontier language model based agents and tabula rasa reinforcement learning algorithms show that these agents cannot solve any of the non-trivial tasks in the BuilderBench. Our analysis throws light on the lack of exploration abilities in these models. 
\looseness=-1
\end{abstract}

\begin{quote}
\emph{Can AI models build a world which today's generative models can only dream of?}
\end{quote}


\begin{center}
  \href{https://rajghugare19.github.io/builderbench/index.html}%
    {\faGlobe\ \textbf{Blog}} \quad
  \href{https://github.com/rajghugare19/builderbench}%
    {\faGithub\ \textbf{Code}}
\end{center}


{\let\thefootnote\relax\footnotetext{$^*$Equal advising.
Correspondence to \texttt{\href{mailto:rg9360@princeton.edu}{rg9360@princeton.edu}}.
}}

\vspace{-1em}
\section{The need for a new benchmark}
\label{sec:intro}
Today's artificial intelligence~(AI) models acquire knowledge by combing through massive collections of human-generated data, enabling them to generate a wide array of images and write a diverse range of stories. While this recipe has been highly successful in domains like vision and language, where models can learn from expert human photographers and writers, it is less clear how to apply this recipe to application areas that humans understand poorly today (e.g., biology, chemistry, architecture)~\citep{ying2025assessingadaptiveworldmodels,  silver2025welcome}. Making progress will require that agents learn not only from human experience, but also from their own, self-collected experience. Agents will have to actively explore and run experiments to extract knowledge about the environment~\citep{Spelke2007core}. Agents will then have to consolidate this knowledge and use it to quickly solve novel tasks. Despite many works recognizing the importance of \textbf{open-ended exploration} and \textbf{learning through experience}~\citep{stanley2017openendedness, adaptiveagentteam2023humantimescaleadaptationopenendedtask}, most benchmarks for building foundation models today focus on learning solely from human data.

This is not for lack of trying. There is a long line of interaction and exploration benchmarks built by researchers in reinforcement learning~(RL)~\citep{ecoffet2021goexplorenewapproachhardexploration, tang2017explorationstudycountbasedexploration}, control~\citep{plappert2018multigoalreinforcementlearningchallenging}, and developmental robotics~\citep{oudeyer2007}, such as maze navigation in ant-maze~\citep{fu2021d4rldatasetsdeepdatadriven}, Montezuma's Revenge~\citep{bellemare_2013atari}, or the handful of tasks in the kitchen environment~\citep{pmlr-v100-gupta20a}. But other than a few exceptions like Minecraft~\citep{guss2019minerllargescaledatasetminecraft}, most widely used benchmarks only allow a handful of diverse behaviors~\citep{rajeswar_2023_urlb, pmlr-v100-gupta20a, fu2021d4rldatasetsdeepdatadriven, tassa2018deepmindcontrolsuite}.
Agents trained on even the most complex of these benchmarks (e.g., StarCraft~\citep{Vinyals2019GrandmasterLI}, AI2Thor~\citep{kolve2022ai2thorinteractive3denvironment}, NetHack~\citep{küttler2020nethacklearningenvironment}) do not seem to learn the same sort of common sense and reasoning skills that agents trained on human text do acquire~\citep{wei2022emergent}.
We argue that the primary reason is that current interactive benchmarks offer limited learning opportunities. Existing benchmarks rarely allow agents to practice a spectrum of skills, ranging from exploration to prediction and from low-level control to high-level reasoning.

We envision a benchmark which enables an open-ended stream of interaction~\citep{pmlr-v235-hughes24a, sigaud2024openendedgcrl}, where training could only ever cover a tiny slice of all possible behaviors.
In the same way that vision models today can paint pictures that go well beyond what is in their training data (e.g., an astronaut mowing the lawn), we desire embodied agentic systems that can solve tasks that go well beyond the tasks they have practiced solving before.
Solving such a benchmark would require agents to have efficient exploration abilities. 
Agents should, in effect, become scientists, performing micro experiments in the environment to discover the laws governing the environment. Once these physical laws have been found, they can be used to make wide-ranging generalizations about how the entire environment works, and how one should act within it. \textit{Our paper constructs an environment where such exploration is possible. One central insight of our paper is to show that this can actually be done using a surprisingly simple setup: block-building.}

\begin{figure}[t]
    \centering
    \vspace{-2em}
    \includegraphics[width=1\linewidth]{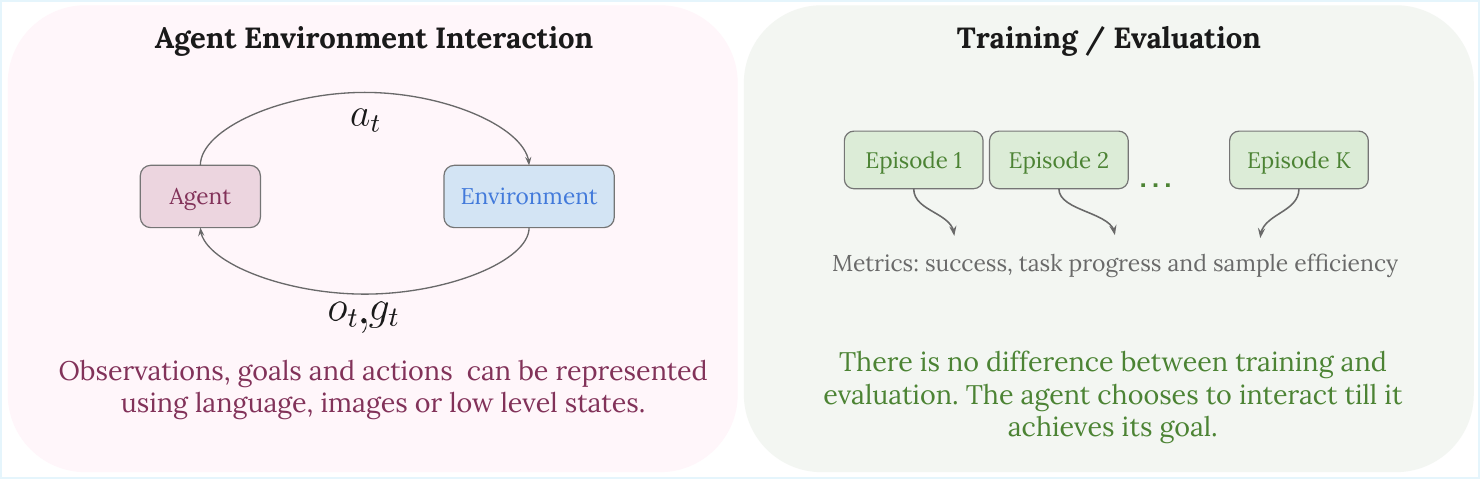}
    \vspace{-1em}    
    \caption{\label{fig:buildstuff-setup} \footnotesize \textbf{The BuilderBench setup.} Vision, language, or tabula rasa agents can be trained / evaluated in BuilderBench. Agents can choose to interact with the simulated environment for multiple episodes to learn to solve the target task. The goal of the agent is to learn to solve \textit{any} given task as efficiently as possible.} 
    \vspace{-1em}
\end{figure}

\textbf{Why block-building?}
Blocks conceptually form an atomic unit, allowing agents to build diverse structures. Many children spend years playing with blocks.
Research in child development highlights that block play builds spatial~\citep{reifel1984block,wexler1998motor, Casey_2008, 10.1093/acprof:oso/9780195304381.001.0001} and arithmetic skills~\citep{Verdine_block_math, cheng2014spatial}. In addition to being useful for early human cognitive development, block-building is a mathematically rich area\footnote{In 2011, \citet{paterson2007maximumoverhang} was awarded the prestigious David P. Robbins Prize in mathematics for improving an upper bound on the maximum overhang using identical blocks.} with a deep history in AI and planning~\citep{GUPTA1992223, ahmad2019formalframeworkrobotconstruction, russel2010}. Building stable structures with blocks requires long-horizon planning and complex reasoning capabilities. While research on reasoning and generalization capabilities has almost become synonymous with large language models in recent years~\citep{touvron2023llamaopenefficientfoundation, deepseekai2025deepseekv3technicalreport, openai2024openaio1card,  geminiteam2025geminifamilyhighlycapable}, block-building allows us to study whether this sort of reasoning and generalization can emerge through exploration and trial-and-error learning. Our central hypothesis, which motivates the use of block-building is that \textit{the space of skills and discoveries that an agent has to know to build all possible structures is so vast, that it is impossible to memorize them at design time.} An agent will always have to interact and make new discoveries to build any given target structure.

\begin{table*}[t]
    \vspace{-2em}
    \centering
    \caption{Qualitative Comparison of BuilerBench with previous benchmarks. BuilderBench is a uniquely positioned robotics benchmark that contains an extensive reasoning task suite and a fast simulator with support for both frontier and tabula rasa RL agents.}
    \label{tab:qaulcom}
    \begin{tabularx}{\textwidth}{
        >{\raggedright\arraybackslash}X
        >{\centering\arraybackslash}m{0.10\textwidth} 
        >{\centering\arraybackslash}m{0.10\textwidth} 
        >{\centering\arraybackslash}m{0.07\textwidth}
        >{\centering\arraybackslash}m{0.12\textwidth}
        >{\centering\arraybackslash}m{0.09\textwidth}
    }
        \toprule
        \textbf{Benchmark} & 
        \textbf{\shortstack{Extensive \\ reasoning \\ task-suite}} &
        \textbf{\shortstack{Frontier \\ model \\ agents }} &
        \textbf{\shortstack{ RL \\ agents }} &
        \textbf{\shortstack{Hardware \\ accelerated}} &
        \textbf{\shortstack{Robotics}} \\
        \midrule
        
        BuilerBench (ours) & \cmark & \cmark & \cmark & \cmark & \cmark\\
        \midrule
        
        ManiSkill~\citep{mu2021maniskillgeneralizablemanipulationskill} & \xmark & \xmark & \cmark  & \cmark & \cmark \\
        \midrule

        FurnitureBench~\citep{heo2023furniturebenchreproduciblerealworldbenchmark} & \xmark & \xmark & \cmark  & \xmark & \cmark\\
        \midrule
        
        BabyAI~\citep{chevalierboisvert2019babyaiplatformstudysample} & \xmark & \xmark & \cmark & \xmark & \xmark\\
        \midrule

        Kinetix~\citep{matthews2025kinetix} & \xmark & \xmark & \cmark & \cmark & \cmark\\
        \midrule

        Craftium~\citep{malagón2025craftiumbridgingflexibilityefficiency} & \xmark & \cmark & \cmark & \xmark & \xmark \\
        \midrule

        NetHack~\citep{küttler2020nethacklearningenvironment} & \xmark & \cmark & \cmark & \xmark & \xmark\\
        
        \bottomrule
    \end{tabularx}
    \vspace{-1em}
\end{table*}

\paragraph{Contributions.}
To this end, we introduce the BuilderBench benchmark. BuilderBench is equipped with a simulated environment, consisting of a robot that can interact with blocks. The environment can output observations in the form of low level state information, images or language descriptions. Agents can control the robot using low level control actions or high level plans to control the robot. This enables many types of agents based on deep RL~\citep{mnih2013atari, pmlr-v80-haarnoja18b}, language models~\citep{brown_gpt2, wei2022emergent}, vision/video language models~\citep{radford2021clip, zhang2023videollamainstructiontunedaudiovisuallanguage} or vision language actions models~\citep{brohan2023rt1roboticstransformerrealworld, intelligence2025pi06vlalearnsexperience} to be evaluated in BuilderBench.

BuilderBench comes equipped with a set of over $50$ challenging tasks~(and growing) which span several orders of magnitude of complexity. Tasks require a wide variety of skills ranging from motor skills like locomotion, grasping and throwing to higher-level skills such as logical reasoning~(commutativity and associativity of pick and place ordering), geometrical reasoning~(maximizing overhangs, packing problems) and intuitive physics~(gravity, friction, toppling, balancing). Tasks also require reasoning about counterweights, buttresses and discovering new tools for performing skills like temporary scaffolding or disassembling.

We evaluate some of the strongest frontier language model based agents available~(at the time of writing) like GPT 5.2, Claude Opus 4.6 and Gemini 3 Flash. We find that these models do not manage to solve most of the hard tasks from the BuilderBench task suite. We provide a detailed qualitative analysis of the failure modes of these models.

Finally, we also open-source single-file implementations of four representative reinforcement learning (RL) algorithms
and three self-supervised data-collection algorithms. Training runs are extremely fast (e.g., training a PPO agent to stack two blocks takes 30 minutes on a single GPU), reducing the barrier to entry for frontier RL research. 

\Cref{tab:qaulcom} provides a qualitative comparison of BuilderBench with other related benchmarks. We provide a detailed comparison of BuilderBench with previous related works in~\Cref{sec:related-work}. Code for the benchmark, all baselines, and experiments is included \href{https://github.com/neurips-submission-2026-abedstwsd99232/BuilderBench}{here}. In the next section, we sketch out what BuilderBench is meant to evaluate and what could be the consequences of a satisfying solution.

\section{What does BuilderBench measure?}
\looseness=-1
BuilderBench measures the exploration efficiency of an agent in an open-ended environment. During training, the goal of an agent is to build a given target structure. The agent can choose to interact with the environment as long as it wants. During this time, it needs to explore various approaches, discover new skills and build the structure through trial and error. We hypothesize that agents which learn to build any unseen target structures efficiently will possess two different type of exploration abilities:

\textbf{Exploration in actual interaction.} Agents need to interact with the environment to gain information about the blocks and the physics. It needs to gain information about how its actions affect the environment. Moreover, it needs to make wide-ranging hypotheses and validate them by trying them out in the environment. Ultimately, learning via interaction and exploration in an open-ended environment\footnote{Especially environments which are not purely lingual.} is a key missing piece in today's AI models. BuilderBench is one way to develop and measure the efficacy of such algorithms based on trial and error.

\textbf{Exploration in thoughts.} There are many tasks in BuilderBench which, despite knowing how the environment works (having a perfect model of the world), are difficult to solve programmatically. The agent needs to explore in the space of "thoughts" to come up with complex and new plans to attempt solving these tasks. The agent does not get any direct feedback on which plans are good, hence we say it has to explore in the space of ``thoughts''. If one were to anthropomorphize further, one would conclude that these tasks require \textit{creative} and \textit{out of the box thinking}.

There are many secondary skills that BuilderBench tasks can also be used to measure.
\textbf{Tool discovery}: Many tasks require the agent to use blocks as tools to provide support or improve reachability. While the latest language models~(at the time of writing) are trained to use tools, BuilderBench requires agents to discover them.
\textbf{Long horizon generalization}: Most tasks can be broken down into sub-tasks whose solutions seem much simpler. Hence, agents need to be able to sequentially compose solutions to sub-tasks.
\textbf{Robotics:} While agents can use high level planners that abstract away the details of control, better robotic systems could potentially be better at solving tasks which require finer control.

\textbf{What it would mean to solve BuilderBench with a satisfying solution}:
The most important and hopeful consequence would be that we would have algorithms which could efficiently learn ``creative'' solutions to unseen problems via interaction with the environment. BuilderBench can also accelerate research in other areas like discovering better interface systems between purely digital agents like LLMs and physical agent like robots. The AI agents might discover new architectural skills~(similar to how chess players learned from alphazero~\cite{silver2017masteringchessshogiselfplay}).

Our central hypothesis, which motivated the use of block-building is that the space of skills and discoveries that an agent has to know to build all possible structures is so vast, that it is impossible to memorize them at design time. In the coming months or years, solutions which attempt to use such memorization might saturate the BuilderBench leaderboard. We are confident that we (or someone else) would be able to come up with new block-building tasks that such an approach will fail to learn to build.

\section{BuilderBench}

The BuilderBench benchmark is equipped with a simulator and a task-suite. The task-suite contains 51 tasks, where each task is a target block structure curated carefully for evaluating unique abilities. The simulator is built using MuJoCo~\citep{todorov2012mujoco}; it consists of a UR5e robot and a Robotiq 2F-85 parallel jaw gripper\footnote{The backbone code for controlling this robot is adapted from~\citet{park2025ogbenchbenchmarkingofflinegoalconditioned}.} interacting with a varying number of blocks. In the following sections, we will describe the environment~(\Cref{sec:environment}), the agent-environment interaction protocol~(\Cref{sec:aei-protocol}), and the task-suite~(\Cref{sec:task-suite}).

\subsection{BuilderBench environment.}
\label{sec:environment}

The environment can be formulated as a Markov decision process~(MDP)~\citep{Sutton1998}, with states $s_t \in \mathcal{S}$ and actions $a_t \in \mathbf{A}$ and transition dynamics $T(s_{t+1} \mid s_t, a_t)$ and a maximum episode length $H$. An additional context parameter $n$ specifies the number of cube-shaped blocks in the environment. Each environment instance contains a single robot that interacts with the $n$ cubes. All interactions approximate real physics simulated using MuJoCo~\citep{todorov2012mujoco}. Additional details about the environment and various potential representations are provided in~\Cref{app:details}.

\textbf{State space.} The low level states include information about the robot and the cubes. This includes the position and velocity of the arm joints and end effector, the gripper opening, the current control inputs of the robot, the position and velocities of all cubes and the current timestep.

\textbf{Observation space.} The simulator exposes all the state information to the agent designer. The agent designer can select which information to provide to the agent and its representation. We provide a language wrapper for the environment which converts the relevant information in a language description of the scene. We also provide functions to render visuals of the scene to enable training vision based models.

\textbf{Action space.} The robot can be manipulated using a $5$ dimensional low level control. These controls consists of delta position and delta yaw of the end effector and delta gripper strength. The language wrapper we provide extends this action space with various high level actions that operate for multiple time-steps such as pick-and-place, pick-and-hold, and end-effector-target. These high level actions allow language model agents to efficiently control the robot. Indeed, language models are able to solve most of the easy tasks from the BuilderBench task-suite using this interface. We do not restrict agent designers from adding more high level actions.

\textbf{Task specification.} Each task corresponds to target structure built using cubes. To specify this structure, we provide a vector of target cube positions~($\mathbb{R}^{3k}$), where $k \leq n$ is the number of cubes in the target cube structure. This allows us to specify target structures that contain fewer cubes than the environment~(see~\Cref{fig:maximum-overhang} for an example).

\subsubsection{Agent-environment interaction Setup}
\label{sec:aei-protocol}
We aim to keep the agent environment interaction loop as general as possible to enable multiple types of agents~(language based, vision based, proprioceptive control; zero shot, in-context learning, RL training). \Cref{fig:buildstuff-setup} provides a visualization of the setup. At the start of the interaction, the agent is provided with the task specification. The agent gets an observation and has to output an action. The agent can choose to interact with the environment for a variable number of episodes~(or until it learns to achieve the goal). The objective of the agent is to learn to build the target structure as fast as possible.

The next section will show how despite this seemingly simple setup, tasks can be arbitrarily complex and long-horizon. Qualitatively, we will see that solving tasks require multiple steps of high-level reasoning.

\subsection{BuilderBench task suite}
\label{sec:task-suite}
In this section, we describe some of the tasks from the BuilderBench task-suite in detail and the design philosophy behind the task-suite. The task-suite is meant to address the challenges highlighted in~\Cref{sec:intro}. We start with a case-study of four different tasks from the BuilderBench task-suite\footnote{We have only visualized the arm of the robot in~\Cref{sec:case-study}.}, which is meant to showcase how each task requires the agent to unlock at least one distinct reasoning ability and compose various high-level skills sequentially. In~\Cref{sec:design-philosophy}, we outline the general design principles that underlie the tasks in the BuilderBench task-suite. The visualizations for all tasks is provided in~\Cref{app:task-suite-table}. 

\subsubsection{A case study of four tasks}
\label{sec:case-study}

\begin{figure}[h]
    \centering
    \vspace{-0.5em}
    \includegraphics[width=1\linewidth]{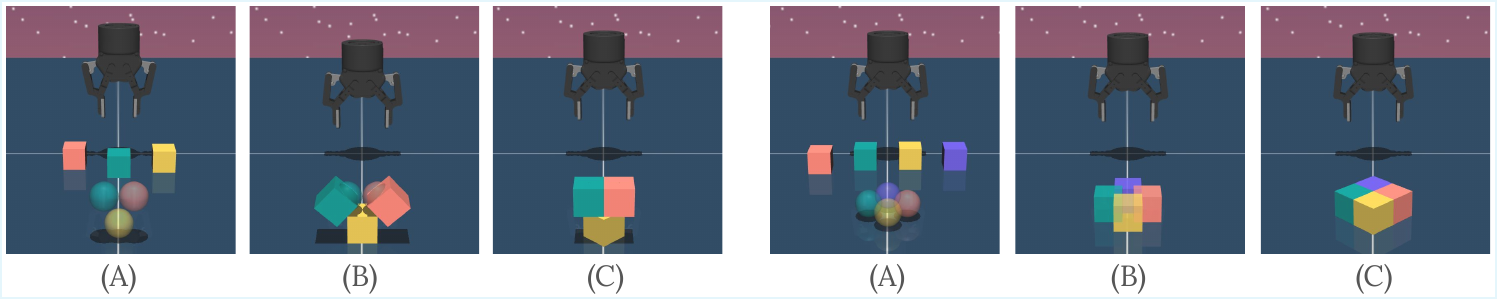}
    \vspace{-1em}
    \caption{ \label{fig:t-block-four-cube-packing} \footnotesize \textbf{T-Block} (Left) and  \textbf{Four Cube Packing} (Right) }
    \vspace{-0.5em}
\end{figure}

\textbf{Example 1: T-Block.} This task requires building a simple T shaped structure with one cube at the base, and two cubes on top~(\Cref{fig:t-block-four-cube-packing}). The second frame (B) shows what many people envision as the solution to this task. However, as shown in the frame, this configuration isn't stable. Solving this task requires the reasoning insight to rotate the bottom cube by about $45^\circ$. Since the diagonal of the cube's top surface is longer than its edge length, the rotated base provides sufficient support for both top cubes, enabling a stable T-shaped structure~(see third frame).
\looseness=-1

\textbf{Example 2: Four Cube Packing.} This task tests geometric reasoning and spatial packing. The target structure is an arrangement of four cube centers placed at some distance along the four cardinal directions on the floor~(see (A) of \Cref{fig:t-block-four-cube-packing}). The distance is chosen such that the placement is impossible with the default cube orientation: the cubes overlap~(see (B)). This results in a packing problem of arranging the cubes such that its centers form the target structure. To solve this, the agent needs to rotate each cube by $45^\circ$ before placing it, which ensures the centers align correctly without collision~(see (C)). Due to the two fingered morphology of the robot, this task cannot be solved using pick and place primitives, but would require nudging the final block in place.
\looseness=-1

\begin{figure}[h]
    \centering
    \includegraphics[width=1\linewidth]{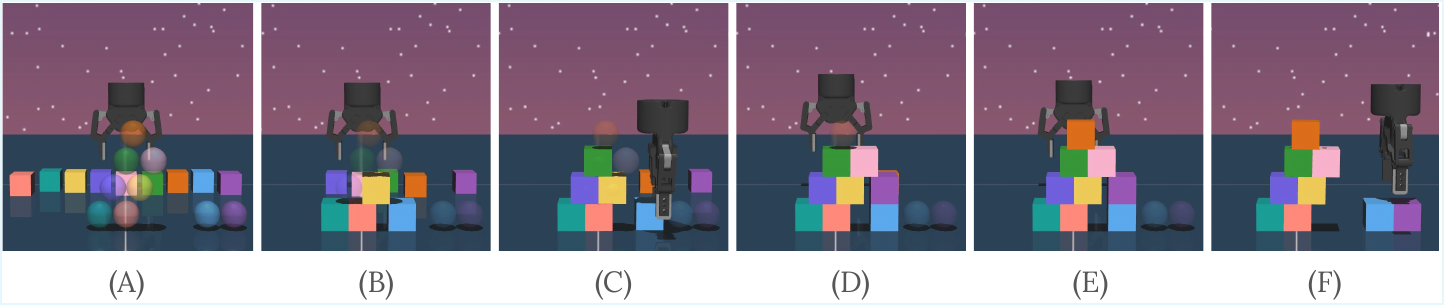}
    \vspace{-1.5em}   
    \caption{\label{fig:learning-tower} \footnotesize \textbf{Leaning Tower}} 
    \vspace{-0.5em}
\end{figure}

\textbf{Example 3: Leaning Tower.} The target is a leaning tower composed of seven blocks and two extra cubes placed on the floor~(see (A) of \Cref{fig:learning-tower}). Solving this task demands building two scaffolds and re-using the first one for the main tower. It also requires an understanding of the concept of counterweights for generating a stable overhang~(an outward extension). The solution itself requires multiple steps of high level planning. After building the base, the yellow block in the second layer must be supported by a temporary scaffold~(see indigo cube in (B)). To stabilize the structure, the agent needs to add counterweights~(the pink and green cubes in (C)) and only then remove the scaffold~(see (D)). To build the third and fourth layer, the agent has to build another set of scaffolds and counterweights. In particular, placing the orange block in the third layer requires a two-cube vertical scaffold~(see (E)). Finally, the tower is completed by adding the counterweights~(the blue and purple cubes in (F)) and removing and repositioning the last scaffold~(see (G)).
\looseness=-1

\begin{wrapfigure}{r}{0.4\textwidth} 
    \centering
    \includegraphics[width=\linewidth]{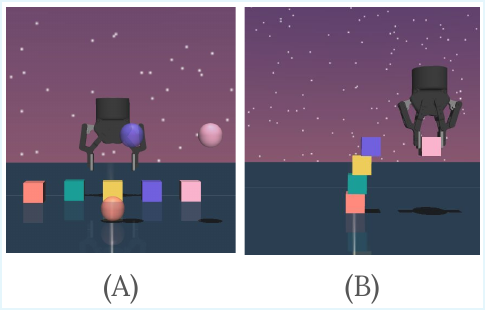}
    \vspace{-1.5em}
    \caption{\footnotesize \textbf{Maximum overhang problem}}
    \label{fig:maximum-overhang}
    \vspace{-0.5em}
\end{wrapfigure}

\textbf{Example 4: Maximum Overhang Problem.} In this task, the environment contains five cubes, but the task only specifies the target positions for three cubes~(see (A) of \Cref{fig:maximum-overhang}). But to put those three cubes in the target location, the agent will need to use all five blocks. To correctly place the green and the yellow cubes~(whose target positions are not specified) in order to complete the task, the agent needs to solve the popular maximum overhang problem~(see~\cite{paterson2007maximumoverhang} for the solution). The main intuition is that at any level, the collective center of the mass of all the cubes above, should not be on the right of the level's boundary. Without such a placement, the task is impossible to solve. The pink cube is specified to ``distract'' the agent from simply holding the indigo cube in place.
\looseness=-1

This case-study illustrates how a block-building setup with a handful of blocks can result in open-ended tasks that can be used to test high-level reasoning abilities. Agents which do not have access to these tasks have almost no chance of encountering them during training. For solving these tasks, we anticipate that agents would have to learn key knowledge priors via exploration~\citep{Spelke2007core}, and learn mechanisms to efficiently use them on the fly~\citep{chollet2019measureintelligence}.
\looseness=-1

\section{Experiments}
\label{sec:experiments}
\looseness=-1
The main goal of our experiments is to ascertain whether the strongest language models~(at the time of writing) display abilities of exploration and learning from interaction to solve BuilderBench tasks. To achieve this, we evaluate three of the strongest available frontier models, GPT 5.2, Claude Opus 4.6, and Gemini 3 Flash (all models at highest levels of reasoning), on all BuilderBench tasks. Because all of these experiments cost money, we have only used one seed per run. We use two different algorithms that enable these models for interaction:

\textbf{Chain of thought~\citep{wei2023chainofthoughtpromptingelicitsreasoning}.} Chain of thought (CoT) reasoning prompts the model to output and store a reasoning chain before outputting every action. All the models we evaluate utilize a significant number of thinking tokens before producing an output. CoT additionally forces the agent to explain its thinking in the output. This is shown to have a significant impact on the performance of language models~\citep{wei2023chainofthoughtpromptingelicitsreasoning}. The CoT agent interacts with the environment for a single episode. Episode lengths have been set so that they provide enough time for exploration and adaptation. The agent has access to at most 16 previous time-steps of observations and actions and its previous reasoning while producing the current output.

\textbf{Reflexion~\citep{shinn2023reflexion}.} Reflexion agents are prompted to maintain a collection of running summaries of their previous experiences and attempts at solving a task. These summaries include running details of the agent's past mistakes, learnings and best plans of action, giving it arbitrary memory. This agent can interact with the environment for multiple episodes~(three in our experiments) and use its past reflections to continually improve its decision-making. The agent has access to at most five previous time-steps of observations and actions, its running summary of the current episode, and summaries of all its previous episodes while producing an output.

Before moving to the results, we describe the language interface used by these agents for interaction.

\textbf{Language interface.} At the start, the agent is provided with an environment system prompt which describes the environment dynamics, semantics of observations and actions, and the expected action schema. See~\Cref{fig:system-prompt} for the exact system prompt. At each timestep of the episode, the interface provides a language description of the scene in a neat tabular format. This description contains the current time, total time and the position and orientation of the end effector. It also contains cube wise positions, orientations, targets and success conditions~(see~\Cref{fig:obs-language} for an example observation). Finally, the agent can control the robot using low level controls or by commanding a high level planner. The planner can be used to move the end effector any pose, pick a cube in different orientations and place or hold a cube in different poses~(see~\Cref{fig:action-language} for an example action).

The task suite contains tasks ranging from very easy to extremely hard. We divided the tasks into two categories~(easy and hard) based on whether they are programmatically straightforward to solve. Tasks which are straightforward compositions of pick and place primitives fall in the easy category.\footnote{Given access to a pick and place planner, stacking 10 blocks shouldn't be considered significantly harder than stacking two blocks.} We evaluate all agents on all the easy~(23/51) and hard tasks~(27/51). The visualizations for all tasks structures and their corresponding difficulty is provided in~\Cref{app:task-suite-table}.

\begin{figure}[t]
    \centering
    \includegraphics[width=1\linewidth]{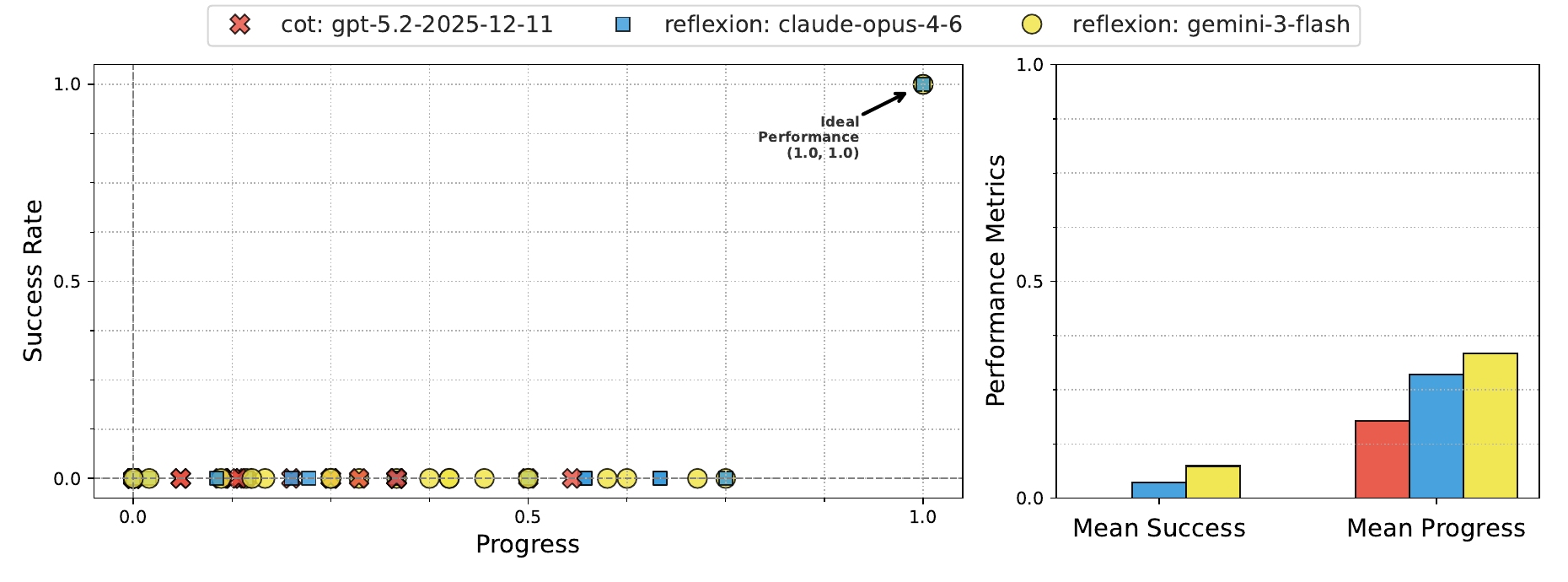}
    \caption{\label{fig:performance-hard} \footnotesize \textbf{Evaluating language model based agents on hard BuilderBench tasks.} Other than two exceptions, all models achieve zero success on all of the 27 hard tasks. The best performing agent, Reflexion with Gemini 3 flash, is able to solve two of the hard tasks from the task-suite. The progress metric is the fraction of blocks the agent manages to place correctly. While the Reflexion agent makes some progress, it does not show an affinity towards making the key discoveries that an agent has to make to solve these tasks.} 
\end{figure}

\subsection{Results}
\label{sec:results}

\begin{figure}[t]
    \centering
    \includegraphics[width=1\linewidth]{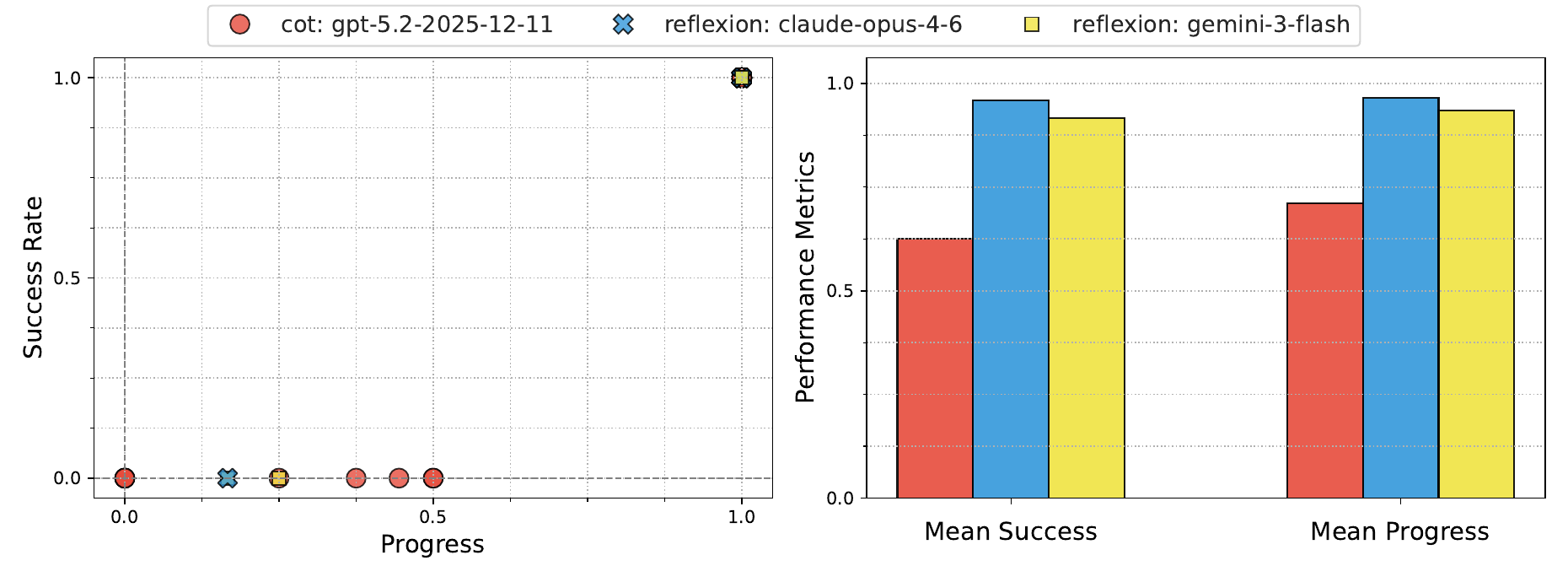}
    \caption{\label{fig:performance-easy} \footnotesize \textbf{Evaluating language model based agents on easy BuilderBench tasks.} The Reflexion agents~\citep{shinn2023reflexion} are able to solve most easy tasks from the BuilderBench task-suite, achieving over 90\% average success across all easy tasks. Most of these tasks are direct applications of the pick and place primitives. Although easy for language models, they still pose a challenge for tabula-rasa RL based agents (see~\Cref{app:tabula-rasa-rl}).} 
\end{figure}

\Cref{fig:performance-hard} shows the results of all agents on the hard tasks. Each point in the scatter plot denotes the performance of an agent solving a task. Success rate~(y axis) is whether the agent is able to build the entire structure or not. Progress~(y axis) is the fraction of blocks the agent manages to place correctly. Other than a single exception, all models achieve zero success on all of the 27 hard tasks. The results on easy tasks are shown in~\Cref{fig:performance-easy}. 

\textbf{Failure modes.} These results suggest that agents are able to compress the observations and extract correct plans and actions for easy tasks. But solving difficult (unseen) tasks requires something beyond mere compression; agents have to show an affinity towards information gathering, make ``out-of-the-box'' hypotheses and validate them through interaction.

By analyzing the reasoning and reflection outputs, and the videos of agents, we observed three primary failure modes of these language model agents:

\begin{itemize}
    \item \textbf{Exploration.} This is the key failure mode which recurs in most tasks where the obvious greedy solution doesn't work. Agents do not showcase hypothesis driven exploration~(can I try balancing... over the other...?), nor do they generate playful hypotheses for information gathering~(let me see what happens if... ). 
    \item \textbf{Planning.} Many times agents try strategies which are clearly going to fail~(trying to place a block where there already is one). Such failure modes should be avoidable if agents have a decent world model of physics and use it to simulate plans. 
    \item \textbf{Fine-grained control.} Agents mostly rely on high level primitives and rarely use skills such as nudging~(although there are some exceptions). This was the most expected failure mode as these models are not trained to explicitly output low level controls.
\end{itemize}

\begin{figure}[t]
    \centering
    \includegraphics[width=1\linewidth]{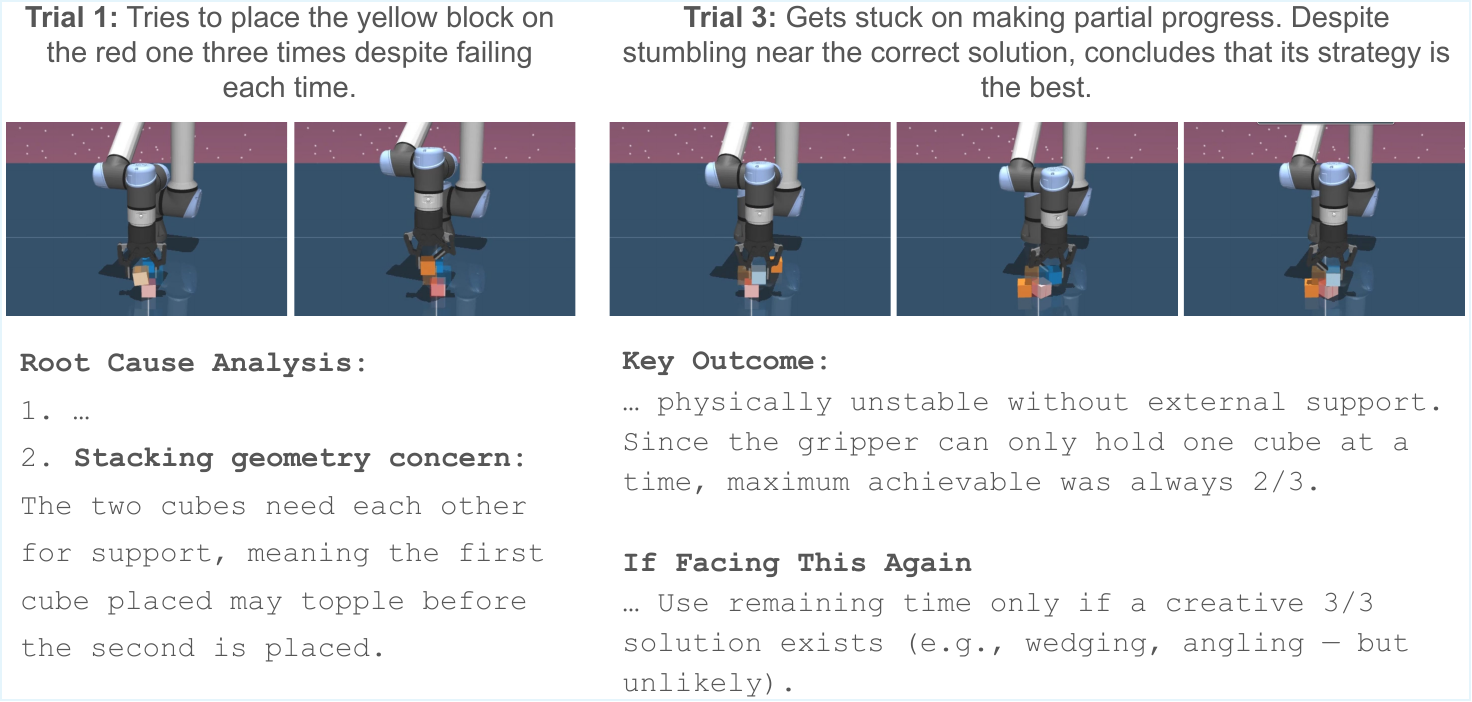}
    \caption{\label{fig:t-block-fail} \footnotesize \textbf{Reflexion agents based on claude opus 4.6 fails on the T-block task~(\Cref{fig:t-block-four-cube-packing}, left).} The text below the images show the intermediate snapshots of summaries maintained by the agent. The agent repeatedly tries the same greedy approach of directly stacking the yellow block at its target despite the structure being unstable. It then summarises that the structure is impossible to build and sticks to holding one of the top blocks in place.} 
\end{figure}

We provide an in-depth analysis of failure modes, including details of agent outputs and videos of agent interaction in a separate \href{https://neurips-submission-2026-abedstwsd99232.github.io/builderbench-blog/}{blogpost}. We visualize one such failure mode of the Reflexion agent trying to solve the T-block task in~\Cref{fig:t-block-fail}. We see that the agent does not explore new strategies and defaults to a greedy straightforward approach despite failing previously.

Finally, we train seven different tabula-rasa RL algorithms to solve the BuilderBench tasks with at most four cubes. We find that as the number of cubes and the complexity of the tasks increase, no algorithm is able to achieve non-zero success. The experimental details and results are presented in~\Cref{app:tabula-rasa-rl} and~\Cref{app:tabula-rasa-ssrl}. 

\section{Limitations and Future work}
\label{sec:limitations}
Despite our hypothesis about building blocks being an open-ended setup, the number of tasks in our benchmark is finite. This is because coming up with new non-trivial tasks is non-trivial. The scope of designing tasks could be drastically improved by adding blocks of new shapes or objects with different properties~(for e.g., magnets). One interesting direction for future work would be to set up an adversarial game between a task designer and a task solver~\citep{openai2021asymmetricselfplayautomaticgoal}.

Another limitation is that methods which memorize the solutions to the tasks in BuilderBench (for e.g., solutions might leak into the training data of future language models) will saturate the scores, but still not provide a satisfying solution. We are confident that we (or someone else) would be able to come up with new block-building tasks that such an approach will fail to learn to build.

Finally, while we use the strongest available language models (at the time of writing) for our evaluation, the agents we evaluate are a tiny subset of all potential approaches. Exploring agents based on vision, control (vision language action models), recursive self improvement or RL fine tuning remain an interesting direction for future work.

\section{Conclusion}
Developing agents that can learn through open-ended exploration and generalize across diverse tasks remains an open problem in AI. Current AI models are pretrained on human generated data. As a result, they largely lack the ability to explore and learn through interaction. We have designed BuilderBench, to accelerate research towards agents that learn via exploration and trial and error. Tasks in BuilderBench are designed to elicit long-horizon planning and reasoning abilities, many implicitly requiring agents to solve problems in physics and mathematics. We evaluated agents based on some of the strongest available language models~(at the time of writing) like  GPT 5.2, Claude Opus 4.6, and Gemini 3 Flash on BuilderBench tasks. Our experiments highlighted that these agents lacked the ability to explore, plan and perform fine grained control. We have open sourced all the code for our experiments. We hope that the research resulting from BuilderBench and our analysis will advance the development of agents that solve problems by interacting with the real world.

\clearpage
\appendix

\begin{figure}[t]
\centering
\begin{tcolorbox}[colback=blue!3, colframe=blue!40, title={\small Environment System Prompt}, fonttitle=\bfseries\small, boxrule=0.5pt, left=4pt, right=4pt, top=2pt, bottom=2pt]
{\ttfamily\scriptsize
You are an agent who must control a UR5e robot arm with a Robotiq 2F-85 parallel jaw gripper in a simulated environment with cube-shaped blocks.\\[2pt]

\textbf{Environment Overview}\\

1. Simulation is implemented using MuJoCo and approximates Newtonian physics.\\
2. All positions are in meters; all rotations are in radians. All coordinates are in the global frame, with z=0 as the ground surface.\\
3. Each cube has an edge length of 0.04 meters.\\
4. The gripper's maximum opening is 0.085 meters.\\
5. The environment provides these observations:\\
    - Current timestep and the total number of timesteps in the episode.\\
    - End-effector position and yaw.\\
    - Potential target position for the end-effector.\\
    - Positions and yaws of all cubes.\\
    - Target locations for some cubes.\\
    - Success condition for cubes with targets.\\
6. You have to output an action (conforming to the defined Action Schema below) at each step with the goal of achieving the success condition for ALL cubes that have an assigned target. Success condition: all cubes are at their respective targets and remain there stably. Move the end effector at its target position (if specified) upon task completion.\\

\textbf{Action Schema}\\

\textbf{Action Types}\\
1. "pick\_and\_place": Executes a Pick -> Lift -> Place -> Retreat plan using low-level controls (no collision avoidance).\\
    - "cube\_id": int — ID of the object to grasp\\
    - "grasp\_yaw": int — 0 or 1 (perpendicular axes for grasping a cube)\\
    - "pos": [x, y, z] — Target position to place the cube\\
    - "yaw": float — Cube placement rotation (radians)\\
    - Note: After placing, the arm retreats to [0.3, 0.0, 0.25].\\

2. "pick\_and\_hold": Executes a Pick -> Lift -> Hold plan (no collision avoidance).\\
    - "cube\_id": int — ID of the object to grasp\\
    - "grasp\_yaw": int — 0 or 1\\
    - "pos": [x, y, z] — Target hold position\\
    - "yaw": float — Target hold rotation (radians)\\
    - Note: The arm holds the cube in the specified pose.\\

3. "eef\_target": Uses a PD controller to move the end-effector to a specified target position and yaw (no collision avoidance).\\
    - "pos": [x, y, z] — Target end-effector position (meters)\\
    - "yaw": float — Target end-effector yaw (radians)\\
    - "gripper": float — 0.0 (open) to 1.0 (closed)\\

4. "low\_level": Applies delta end-effector control for a few timesteps (fine-grained control; plan sequences to achieve high-level tasks).\\
    - "action": [delta\_x, delta\_y, delta\_z, delta\_yaw, delta\_gripper\_strength]\\

\textbf{Output Format}\\
Always output actions as a single, valid JSON object with the specified key/value structure.\\

\textbf{Examples}\\
- Pick and Place:\\
  {"type": "pick\_and\_place", "cube\_id": 0, "grasp\_yaw": 0, "pos": [0.5, -0.2, 0.02], "yaw": 1.57}\\

- Pick and Hold:\\
  {"type": "pick\_and\_hold", "cube\_id": 0, "grasp\_yaw": 0, "pos": [0.5, -0.2, 0.2], "yaw": 0.0}\\

- End-Effector Target:\\
  {"type": "eef\_target", "pos": [0.45, 0.1, 0.3], "yaw": 1.57, "gripper": 1.0}\\

- Low Level:\\
  {"type": "low\_level", "action": [0.3, 0.0, 0.1, 0.0, 1.0]}\\
}
\end{tcolorbox}
\captionof{figure}{Environment system prompt provided to language model agents.}
\label{fig:system-prompt}
\end{figure}

\begin{figure}[t]
\centering
\begin{tcolorbox}[colback=gray!5, colframe=gray!60, title={\small Example Language Observation}, fonttitle=\bfseries\small, boxrule=0.5pt, left=4pt, right=4pt, top=2pt, bottom=2pt]
{\ttfamily\scriptsize
Time: 2.88 / 18.00\\

End Effector State\\
- End Effector: pos=[0.323, 0.004, 0.228], yaw=-0.032, target=[0.300, 0.000, 0.250]\\
- Gripper: 0.000\\

Cube State\\
- Cube 0: pos=[0.452, -0.001, 0.020], yaw=0.009, target=[0.450, 0.000, 0.020], success=True\\
- Cube 1: pos=[0.300, 0.000, 0.020], yaw=0.000, target=[0.450, 0.000, 0.060], success=False\\
- Cube 2: pos=[0.300, 0.080, 0.020], yaw=0.000, target=[0.450, 0.000, 0.100], success=False\\
}
\end{tcolorbox}
\captionof{figure}{Example language observation provided by the language interface of BuilderBench for a cube stacking task. The observation contains the current episode time, the total episode time, the state of the end effector and the state and success conditions for each cube.}
\label{fig:obs-language}
\end{figure}

\begin{figure}[t]
\centering
\begin{tcolorbox}[colback=red!3, colframe=red!40, title={\small Example Language Action}, fonttitle=\bfseries\small, boxrule=0.5pt, left=4pt, right=4pt, top=2pt, bottom=2pt]
{\ttfamily\scriptsize
"{'type': 'pick\_and\_place', 'cube\_id': 1, 'grasp\_yaw': 0, 'pos': [0.45, 0.0, 0.06], 'yaw': 0.0}"
}
\end{tcolorbox}
\captionof{figure}{Example of an action output by an LLM agent in the desired JSON format. The agent commands the environment to pick cube $1$ and place in a particular pose.}
\label{fig:action-language}
\end{figure}

\section{Related Works}
\label{sec:related-work}
AI benchmarks have driven progress in the field. Benchmarks such as MNIST~\citep{mnist}, ImageNet~\citep{russakovsky2015imagenetlargescalevisual}, Atari~\citep{bellemare_2013atari}, Gym~\citep{brockman2016openaigym}, WMT~\citep{chelba2014billionwordbenchmarkmeasuring}, SWE-bench~\citep{jimenez2024swebenchlanguagemodelsresolve}, ARC-AGI~\citep{chollet2019measureintelligence} have propelled research in deep learning, vision, RL and natural language processing. The aim of BuilderBench is to similarly propel research on agents that learn through trial and error. Below we discuss various aspects of this problem and prior attempts to tackle and benchmark them.

Reinforcement learning (RL) studies agents that learn through interaction. Standard RL benchmarks~\citep{bellemare_2013atari, brockman2016openaigym, tassa2018deepmindcontrolsuite, crafter, küttler2020nethacklearningenvironment, koyamada2023pgx, bonnet2024jumanji} have agents learn to maximize hand-designed rewards to solve a task of interest. These environments require agents to extract their own knowledge and novel solutions~(e.g., endlessly bouncing the ball in breakout from DQN~\citep{mnih2013atari} or the famous ``Move 37'' from AlphaGo~\citep{Silver_2016_alphago}). However, these environments focus on solving a small range of tasks. As a result, RL agents typically possess narrow or poor generalization capabilities~\citep{kirk_2023gen}. Recently, such environments have also been adapted to evaluate language model or vision model based agents~\cite{stojanovski2025reasoninggymreasoningenvironments, paglieri2025balrogbenchmarkingagenticllm, guertler2025textarena}. But ultimately the adaptations face the same problem. The type of generalization that is desired is not just towards perturbed observations or dynamics~\citep{stone2021distractingcontrolsuite, cobbe2020leveragingproceduralgenerationbenchmark}, but towards learning to solve diverse unseen tasks~\citep{epistemic_pomdp}. 

Unsupervised RL is centered on devising objectives that let agents learn through trial and error without any rewards. Such methods usually try to learn generally useful skills~\citep{Gregor2016VariationalIC, eysenbach2018diversity} or collect exploratory data~\citep{lee2020smm, tang2017explorationstudycountbasedexploration, osband2016deepexplorationbootstrappeddqn}. But it is not clear how scalable these objectives are, mainly because the standard unsupervised RL benchmarks~\citep{rajeswar_2023_urlb, fu2021d4rldatasetsdeepdatadriven, tassa2018deepmindcontrolsuite} contain only a handful of similar downstream tasks for evaluation. Hence, to properly evaluate generalization properties of agents, benchmarks need to permit sufficiently complex and open ended interaction. 

Another set of methods that are closely related are ones which treat the problem of efficiently generalizing to unseen tasks as a learning problem itself. Meta-learning~\citep{multitask_learning, finn2017modelagnosticmetalearningfastadaptation, schmidhuber:1987:srl} and few-shot learning~\citep{matching_networks, protonet} fall under this category. Initial progress was driven by benchmarks that arranged common supervised learning tasks episodically, testing how quickly models adapt to new tasks~\citep{omniglot, dhillon2020baselinefewshotimageclassification}. Later work found that self-supervised pre-training on diverse datasets provided enough prior knowledge to directly solve most of the common supervised learning tasks~\citep{radford2021clip, brown_gpt2, devlin2018bert}, blurring the boundary between memorizing prior knowledge and efficiently generalizing. We argue that open-ended domains and tasks are needed to disentangle the two. ARC-AGI~\citep{chollet2019measureintelligence} uses the open-ended domain of discrete puzzles to measure a model's ability to efficiently use its priors. ARC tests models on a set of novel puzzles that require on-the-fly composition of a minimal set of core principles~\citep{chollet2019measureintelligence, Spelke2007core}. BuilderBench is similarly structured. Solving tasks from the BuilderBench task-suite not only requires a concrete set of priors (e.g., an understanding of Newtonian physics), but requires using these priors to build unseen structures on-the-fly. Unlike ARC-AGI, where priors are directly provided through examples of solved puzzles, in BuilderBench agents have to discover priors on their own through interaction.

In addition to exploration and generalization, the BuilderBench task-suite highlights how block-building can also be used to evaluate various types of reasoning abilities~(see~\Cref{sec:case-study} for details). Many of these abilities are typically studied only in isolation. For e.g., intuitive physics is evaluated in~\citet{chow2025physbench, riochet2020intphysframeworkbenchmarkvisual}, motor skills in~\citet{james2019rlbenchrobotlearningbenchmark, Melnik2021UsingTS}, planning in~\citet{valmeekam2023planbenchextensiblebenchmarkevaluating}, mathematical reasoning in~\citet{lewkowycz2022solving, ahn2024largelanguagemodelsmathematical}. In recent years, reasoning is almost exclusively studied using language models pretrained on data. However, BuilderBench allows us to evaluate and visualize reasoning that is not grounded in language and not learned using human data.

There exists other similar benchmarks which aim to capture open-ended interaction like Kinetix~\citep{matthews2025kinetix}, XLand~\citep{openendedlearningteam2021openendedlearningleadsgenerally}, and Minecraft~\citep{guss2019minerllargescaledatasetminecraft}. Kinetix provides a diverse set of rigid body tasks, constrained to 2D, to test zero shot generalization of agents. Tasks in Kinetix are procedurally generated. Unlike BuilderBench, these tasks do not clearly test diverse logical and mathematical reasoning abilities. XLand provides a vast set of multi-agent video-game like tasks, but is closed source and not readily available for academic research. Minecraft is a popular open-ended game that revolves around building various artifacts with blocks that has been used to develop generalist agents from scratch~\citep{hafner2024masteringdiversedomainsworld,Ma2022VIPTU, zhao2024thinkembodiedagentvirtual, malagon2025craftiumbridgingflexibilityefficiency}. While based on the similar block-building foundations and an appealing benchmark, we believe BuilderBench is better suited for academic research due to its simplicity, the much faster speed of its simulator and an extensive carefully curated task-suite. Finally, BuilderBench is fully open source, making all of its components flexible and easy to adapt.

\section{Environment Details}
\label{app:details}
\Cref{fig:system-prompt} shows the system prompt provided to language model agents, which describes the environment and the action schema. \Cref{fig:obs-language} and \Cref{fig:action-language} provide an example of a language based observation and a language based action.

The entire essential information about the scene can be conveyed by representing the state of the robot~(joint angles and velocities, end effector poses and velocities) and the state of the blocks~(poses and velocities). This information can be represented and compressed in language using the provided language wrapper. Users can also render multiple images of the scene via various cameras using our code.

\begin{figure}[t]
    \centering
    \includegraphics[width=1\linewidth]{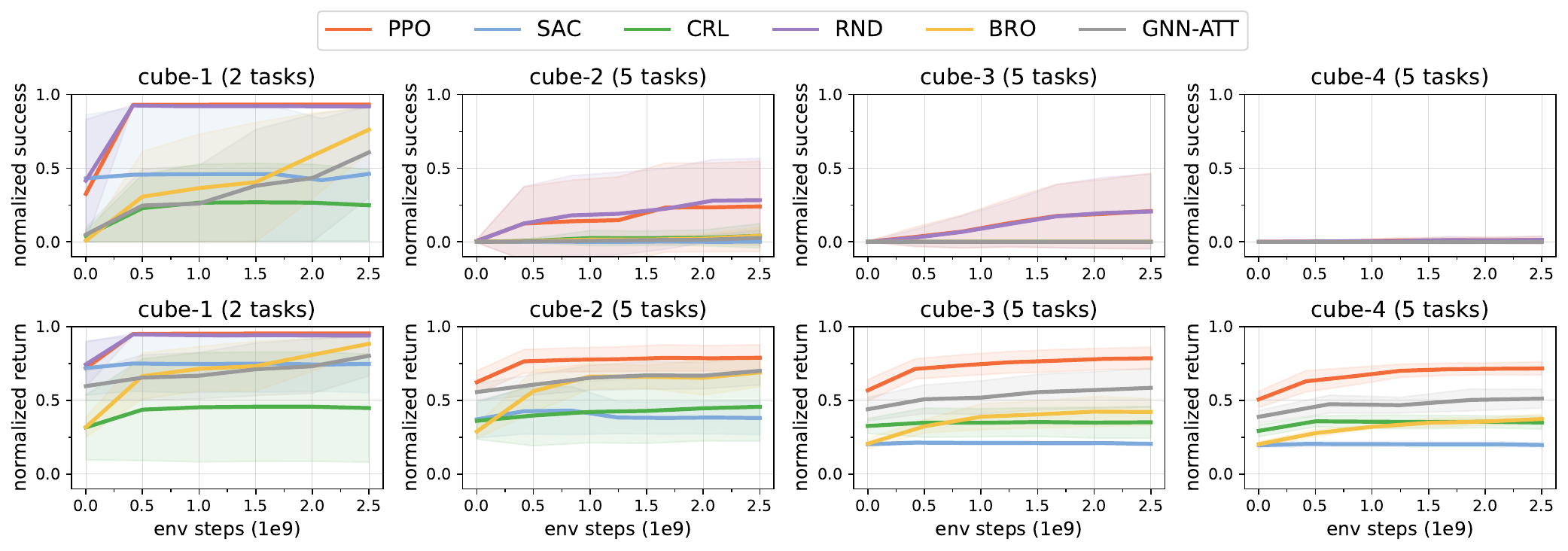}
    \caption{\label{fig:results-srl} \footnotesize \textbf{RL from scratch.} As the number of cubes and the complexity of the tasks increase, no RL algorithm is able to achieve a non zero success.}
\end{figure}

\begin{figure}[t]
    \centering
    \includegraphics[width=1\linewidth]{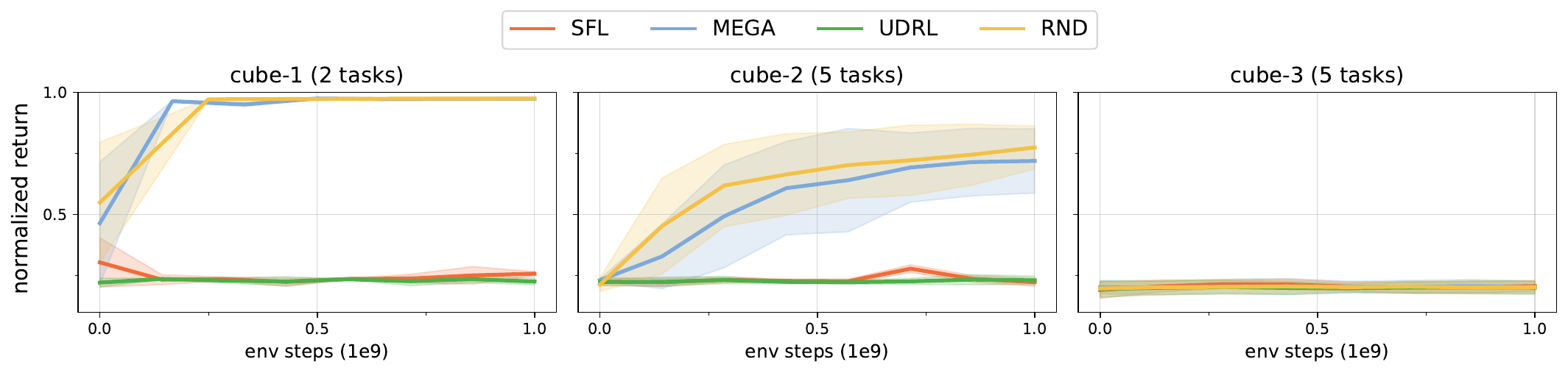}
    \vspace{-1em}    
    \caption{\label{fig:results-ssrl} \footnotesize \textbf{Self-supervised evaluation on BuilderBench task-suite.} We evaluate MEGA~\citep{mega_2020} and SFL~\citep{rutherford2024no} on 12 of the lowest complexity (yet still difficult) tasks from our task suite. The results show that directly using these algorithms out of the box only succeeds for the simplest tasks.
    }
\end{figure}

\section{Design philosophy behind the BuilderBench task-suite.}
\label{sec:design-philosophy}
The primary goal of the task-suite is to capture the main challenges in evaluating open-ended exploration and generalization~(highlighted in \Cref{sec:intro}) and provide a meaningful feedback signal for algorithmic research. To best support these goals, we followed the following design principles:

\textbf{Solving different tasks should require distinct skills.} For example, once an agent learns how to pick and place two blocks, extending this to three or more independent blocks does not qualitatively require an additional ability. We have designed tasks such that they demand a range of motor skills, including grasping, nudging, and throwing. Importantly, tasks also require logical reasoning skills, such as commutativity and associativity of blocks (pick and place ordering), induction (stacking $n$ blocks vs stacking $n+1$ blocks), geometry, and intuitive physics. 

\textbf{Most tasks should be solvable by humans.} To ensure that solving the tasks is theoretically possible, we manually solved most tasks using the same action space as the agent. We also provide scripts that allow researchers to explore the environments and attempt to solve tasks themselves\footnote{Please checkout our code for the script to teleoperate the robot.}.

\textbf{Tasks should range from very easy to extremely hard.} This is an important feature of BuilderBench, meant to provide breadcrumbs of feedback to go from current algorithms capable of solving only the easy tasks to agents that can build anything.

\textbf{Tasks should include some whose solutions are unknown even to the authors.} One aim of BuilderBench is also to see if artificial agents can come up with solutions to problems whose solutions are unknown. Hence, we have included a small minority of structures which even we were not able to build.

\section{Reinforcement learning from scratch}
\label{app:tabula-rasa-rl}
This section provides details on our RL experiments. These experiments are conducted with a simpler setup as compared to our language model experiments. In the simulator, instead of the entire robot, we only include the robot gripper which can move around in 3-D space. We do not include the entire robot to make the simulator faster and parallelizable with jax's jit compilation. Moreover, inverse kinematics is a solvable and orthogonal problem.

We benchmark six RL algorithms, proximal policy optimization~(\textbf{PPO})~\citep{schulman2017proximalpolicyoptimizationalgorithms}, soft actor critic~(\textbf{SAC})~\citep{pmlr-v80-haarnoja18b}, contrastive RL~(\textbf{CRL})~\citep{eysenbach2022contrastive}, random network distillation~(\textbf{RND})~\citep{burda2018exploration}, bigger-regularized-optimistic~(\textbf{BRO})~\citep{nauman2024biggerregularizedoptimisticscaling} and graph-attention-network~(\textbf{GNN-ATT})~\citep{ghasemipour2022blocksassemblelearningassemble}. 

\textbf{Reward function.} These experiments use dense rewards which are permutation invariant to the cube order. Cubes are said to form the target structure if the distance between each cube and its corresponding target is less than two centimeters. At each timestep, every cube is assigned a specific target position from the target structure. This assignment is calculated such that the total sum of distances between the cubes and their assigned targets is minimized. This is a convex optimization problem and can be solved efficiently with GPUs using the Hungarian algorithm implemented in jax~\citep{jax2018github}. The dense rewards are calculated by applying $1 - \tanh(x)$ to the best assigned distances and summing them over all cubes. As distances tend to zero, the reward tends to N~(number of cubes in the environment).

\textbf{Results.} The benchmarking results on tasks with atmost 4 cubes are provided in~\Cref{fig:results-srl}. As the number of cubes and the complexity of the tasks increase, no RL algorithm is able to achieve a non zero success. Issues like sample inefficiency and exploration are the main bottleneck for RL algorithms. We leave a more detailed study of failure modes and improvements for tabula rasa RL algorithms for future work.

\section{Self-supervised RL pretraining}
\label{app:tabula-rasa-ssrl}
To evaluate exploration based RL pretraining, we propose a different protocol. The agent first interacts with the environment, but does not receive any task specification during training. The agent's goal is to explore its environment to acquire general knowledge and skills that might help it to solve future tasks. The agent has to learn a task conditioned policy~\citep{Kaelbling1993LearningTA}, which can take as input a state ($\mathbb{R}^{11 + 13n}$) as well as a task specification ($\mathbb{R}^{3k}$). Each environment has a number of hand-designed tasks associated with it~(\Cref{app:task-suite-table}). The agent is evaluated by running its task-conditioned policy on these tasks and measuring the reward obtained by it.

During training, it is highly unlikely that the agents will have seen these hand-designed tasks. Hence, to solve this protocol, agents will have to learn general reusable skills and concepts through purely self-supervised interaction. Many of these tasks are very difficult and unsolvable by the initial algorithms we tried. 

We implemented four algorithms, sampling for learnability~(\textbf{SFL}~\citep{rutherford2024no}), maximum entropy gain exploration~(\textbf{MEGA}~\citep{mega_2020}), upside down RL~(\textbf{UDRL})~\citep{schmidhuber2020reinforcementlearningupsidedown} and random network distillation~(\textbf{RND})~\citep{burda2018explorationrandomnetworkdistillation}. SFL and MEGA sample autotelic goals from previously visited states, for the agent to learn to reach them. SFL is an unsupervised environment design~\citep{ued_dennis} algorithm, which samples goals with the highest learnability~(variance of success). MEGA is an unsupervised goal sampling~\citep{florensa2018automaticgoalgenerationreinforcement, openai2021asymmetricselfplayautomaticgoal} algorithm, which samples goals inversely proportional to their visitation density. Both algorithms are implemented using proximal policy optimization~(PPO)~\citep{schulman2017proximalpolicyoptimizationalgorithms}.UDRL and RND are self-supervised algorithms. UDRL learns to reach previously explored goals using hindsight relabelling~\citep{andrychowicz2018hindsightexperiencereplay} and RND explores the environment using an intrinsic reward bonus. Both of these algorithms sample data collection goals using MEGA. All algorithms are trained in environments with one, two and three cubes and the learned policies are tested on the respective tasks from the task-suite~(\Cref{app:task-suite-table}) at various points during training. We report normalized episodic success and returns in~\Cref{fig:results-ssrl}.

\paragraph{Results.} As seen in~\Cref{fig:results-ssrl}, both algorithms achieve trivial performance on tasks with three cubes. MEGA is able to complete both tasks with one cube, and shows improvement on tasks with two cubes. While these results indicate that the tested algorithms are not directly scalable to complex tasks, it primarily underscores the inherent difficulty of the task setup itself. We believe that research in developing new algorithms~(or revisiting old ones) is required to solve these tasks.

\section{Compute Usage}
\label{app:compute-usage}
The LLM based experiments in~\cref{fig:performance-easy} and~\cref{fig:performance-hard} did not require any GPUs. Some runs from these figures were replicated on a relatively low end personal laptop with AMD Ryzen 5 4600H CPU. The experiments in~\cref{fig:results-srl} and~\cref{fig:results-ssrl} were performaned using A100 and H100 GPUs. Each run can be performed on a single A100 or H100 GPU with atleast 40 GB of VRAM.

\clearpage
\section{BuilderBench Task Suite}
\label{app:task-suite-table}

\Cref{app:task-suite-table} provides the visualizations for the target structures of each tasks along with its assigned difficulty level.

\begin{longtable}{ c c c }
\toprule
\textbf{Task Name} & \textbf{Image} & \textbf{Difficulty} \\
\midrule
\endfirsthead

\toprule
\textbf{Task Name} & \textbf{Image} & \textbf{Difficulty} \\
\midrule
\endhead

\bottomrule
\endfoot

\bottomrule
\endlastfoot

cube-1-task-1 & \includegraphics[width=4cm, margin=0pt 6pt 0pt 6pt, valign=c]{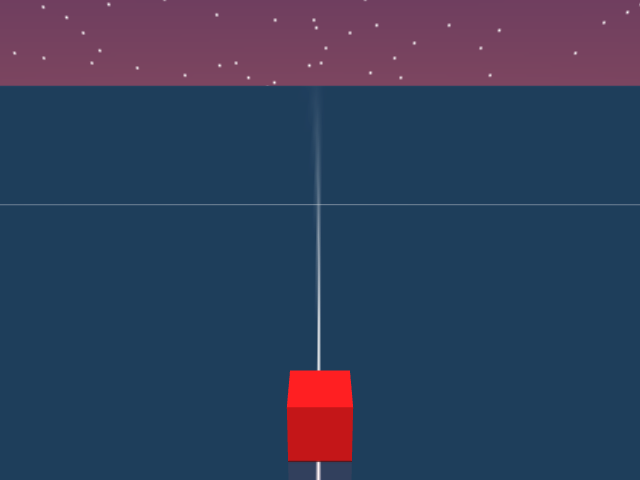} & Easy \\ \midrule
cube-1-task-2 & \includegraphics[width=4cm, margin=0pt 6pt 0pt 6pt, valign=c]{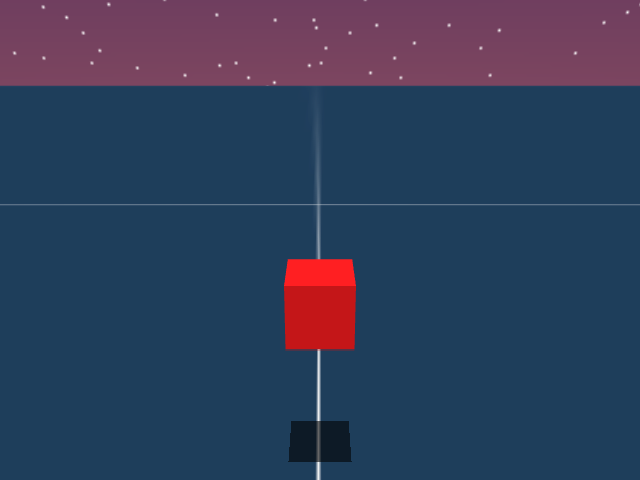} & Easy \\ \midrule

cube-2-task-1 & \includegraphics[width=4cm, margin=0pt 6pt 0pt 6pt, valign=c]{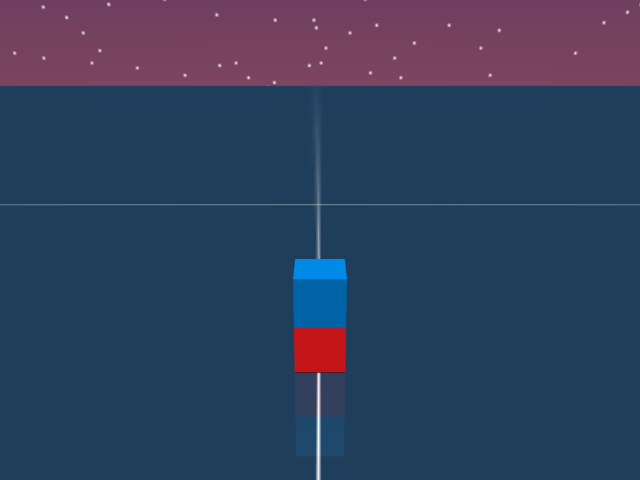} & Easy \\ \midrule
cube-2-task-2 & \includegraphics[width=4cm, margin=0pt 6pt 0pt 6pt, valign=c]{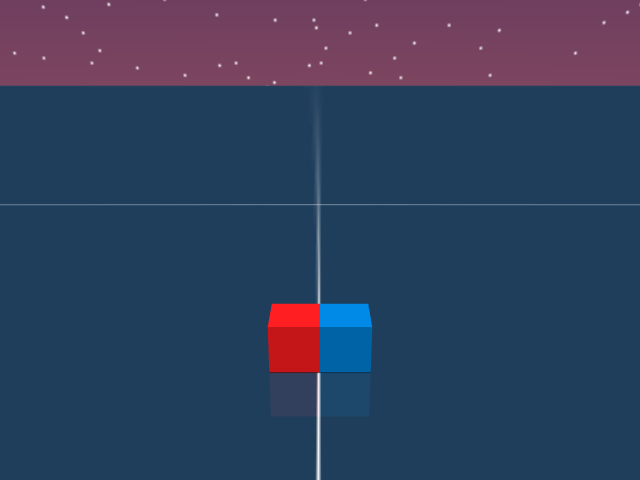} & Easy \\ \midrule
cube-2-task-3 & \includegraphics[width=4cm, margin=0pt 6pt 0pt 6pt, valign=c]{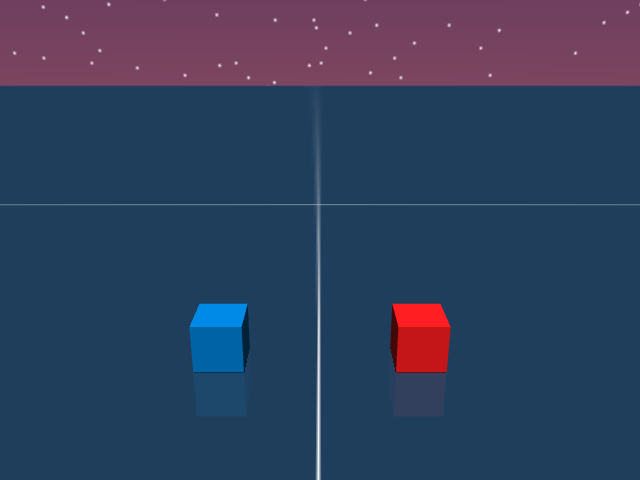} & Easy \\ \midrule
cube-2-task-4 & \includegraphics[width=4cm, margin=0pt 6pt 0pt 6pt, valign=c]{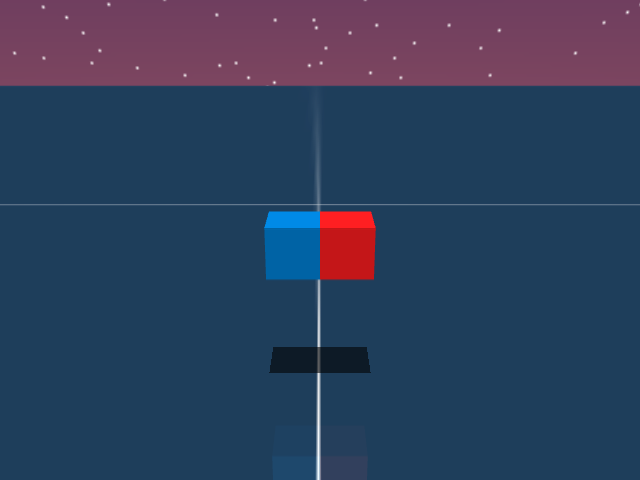} & Hard \\ \midrule
cube-2-task-5 & \includegraphics[width=4cm, margin=0pt 6pt 0pt 6pt, valign=c]{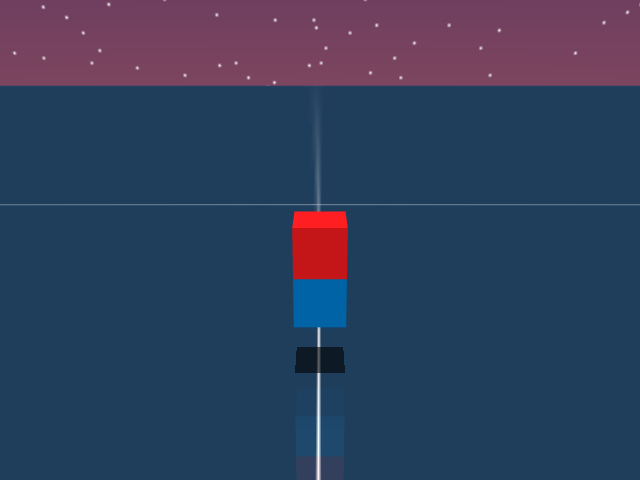} & Hard \\ \midrule

cube-3-task-1 & \includegraphics[width=4cm, margin=0pt 6pt 0pt 6pt, valign=c]{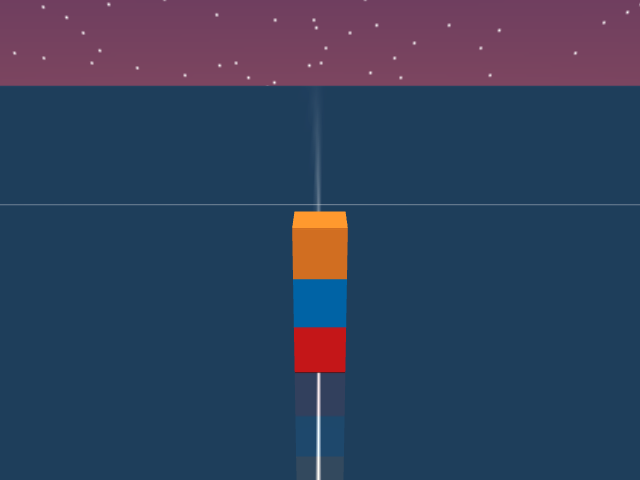} & Easy \\ \midrule
cube-3-task-2 & \includegraphics[width=4cm, margin=0pt 6pt 0pt 6pt, valign=c]{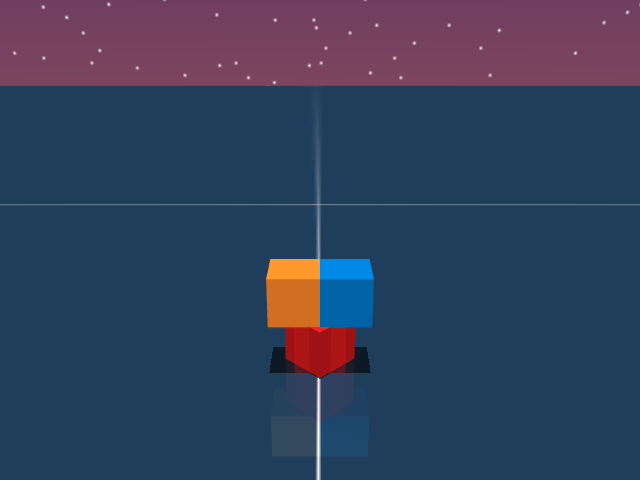} & Hard \\ \midrule
cube-3-task-3 & \includegraphics[width=4cm, margin=0pt 6pt 0pt 6pt, valign=c]{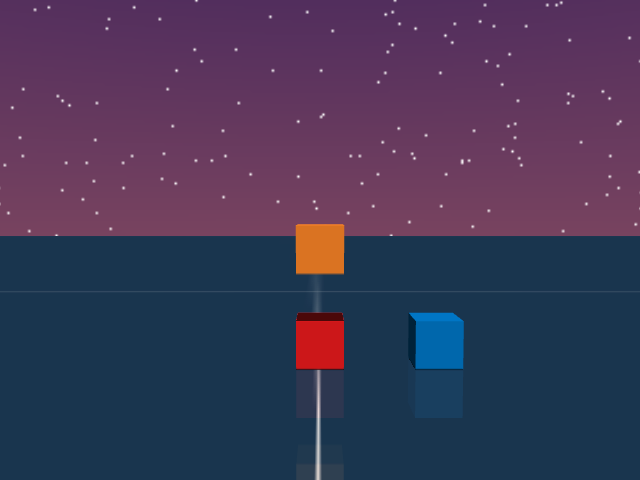} & Easy \\ \midrule
cube-3-task-4 & \includegraphics[width=4cm, margin=0pt 6pt 0pt 6pt, valign=c]{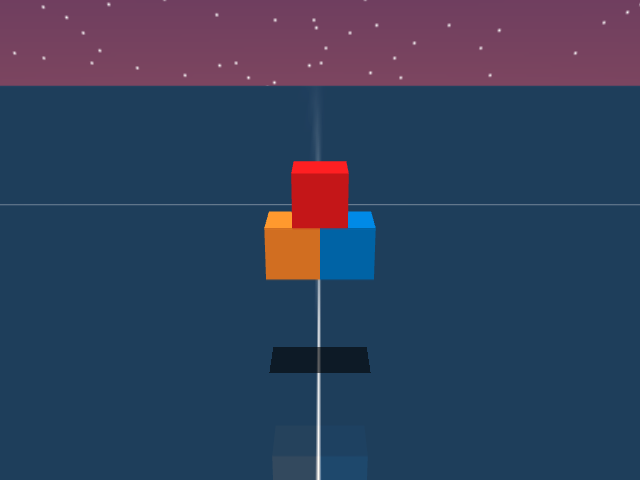} & Hard \\ \midrule
cube-3-task-5 & \includegraphics[width=4cm, margin=0pt 6pt 0pt 6pt, valign=c]{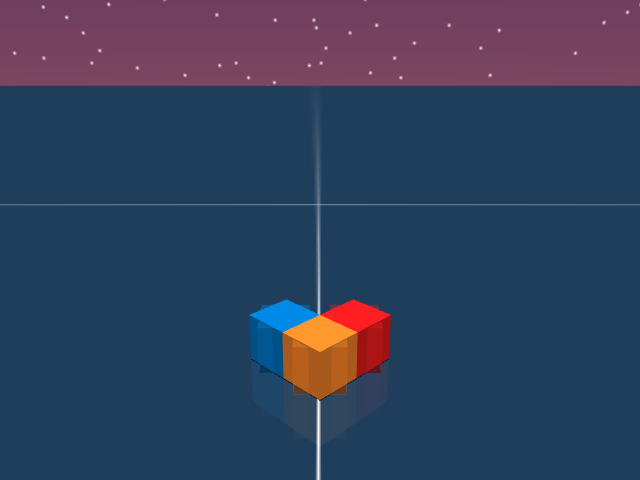} & Hard \\ \midrule
cube-3-task-6 & \includegraphics[width=4cm, margin=0pt 6pt 0pt 6pt, valign=c]{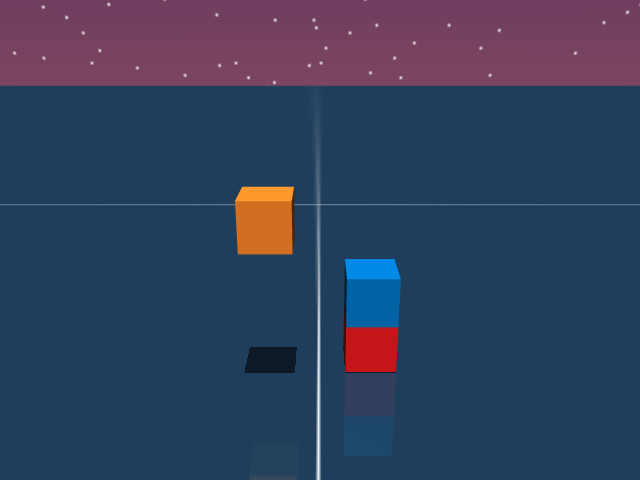} & Easy \\ \midrule

cube-4-task-1 & \includegraphics[width=4cm, margin=0pt 6pt 0pt 6pt, valign=c]{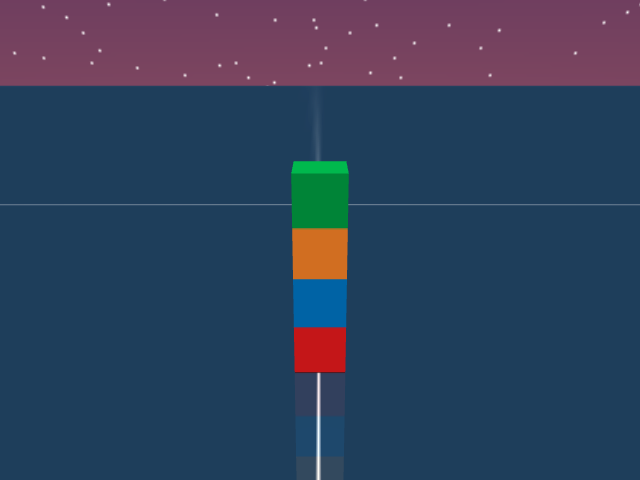} & Easy \\ \midrule
cube-4-task-2 & \includegraphics[width=4cm, margin=0pt 6pt 0pt 6pt, valign=c]{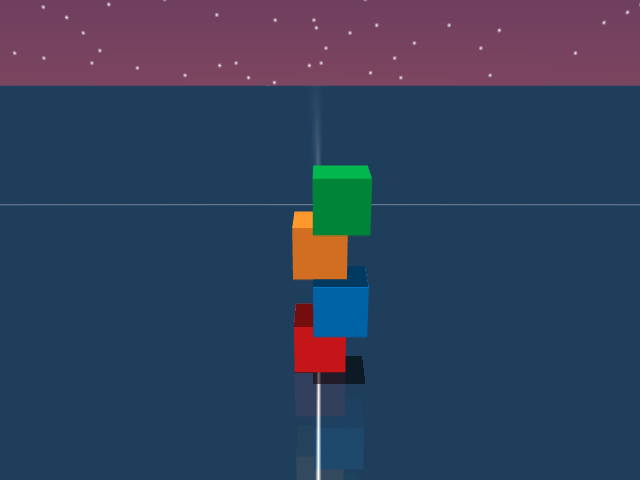} & Easy \\ \midrule
cube-4-task-3 & \includegraphics[width=4cm, margin=0pt 6pt 0pt 6pt, valign=c]{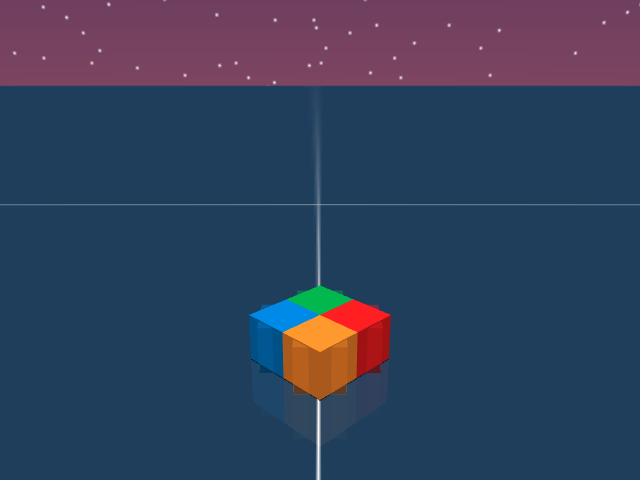} & Hard \\ \midrule
cube-4-task-4 & \includegraphics[width=4cm, margin=0pt 6pt 0pt 6pt, valign=c]{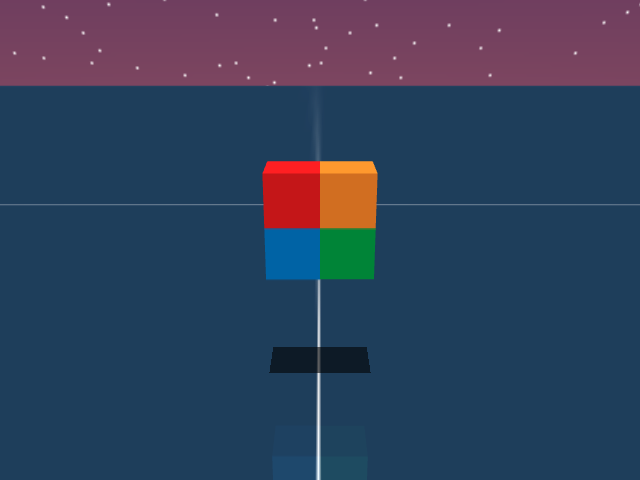} & Hard \\ \midrule
cube-4-task-5 & \includegraphics[width=4cm, margin=0pt 6pt 0pt 6pt, valign=c]{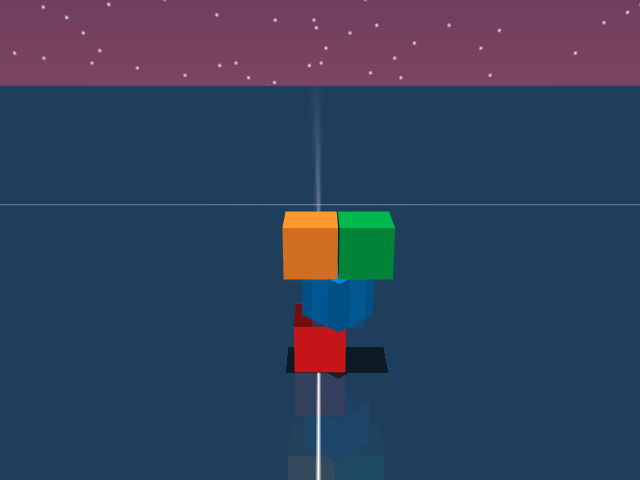} & Hard \\ \midrule
cube-4-task-6 & \includegraphics[width=4cm, margin=0pt 6pt 0pt 6pt, valign=c]{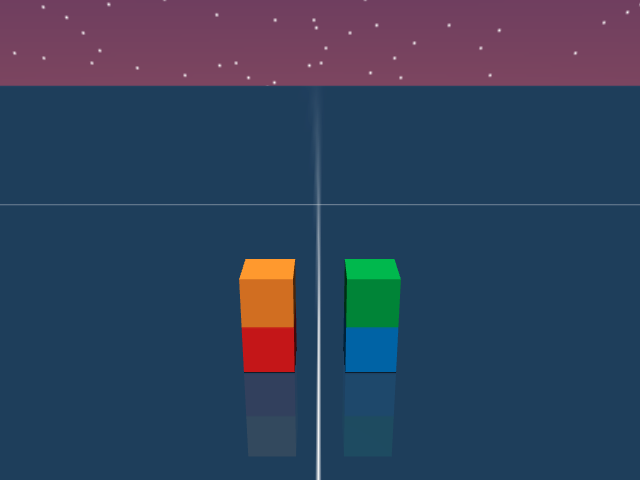} & Easy \\ \midrule

cube-5-task-1 & \includegraphics[width=4cm, margin=0pt 6pt 0pt 6pt, valign=c]{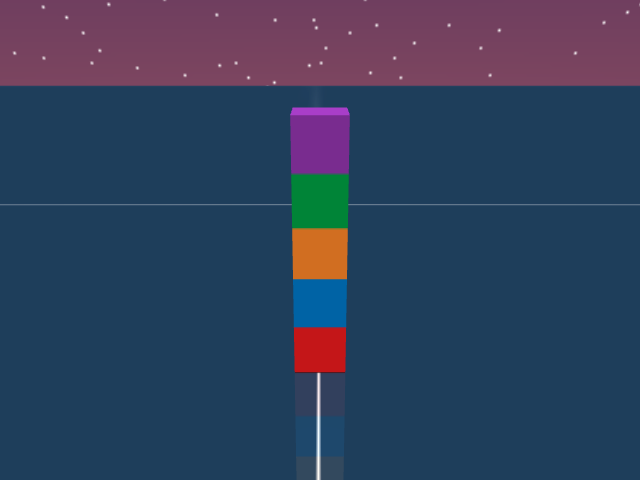} & Easy \\ \midrule
cube-5-task-2 & \includegraphics[width=4cm, margin=0pt 6pt 0pt 6pt, valign=c]{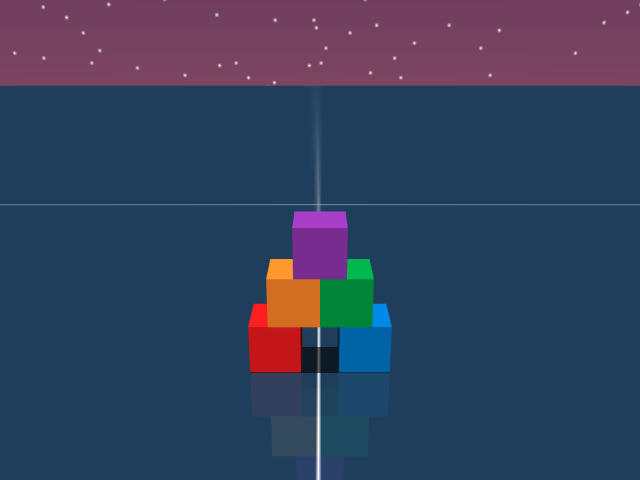} & Easy \\ \midrule
cube-5-task-3 & \includegraphics[width=4cm, margin=0pt 6pt 0pt 6pt, valign=c]{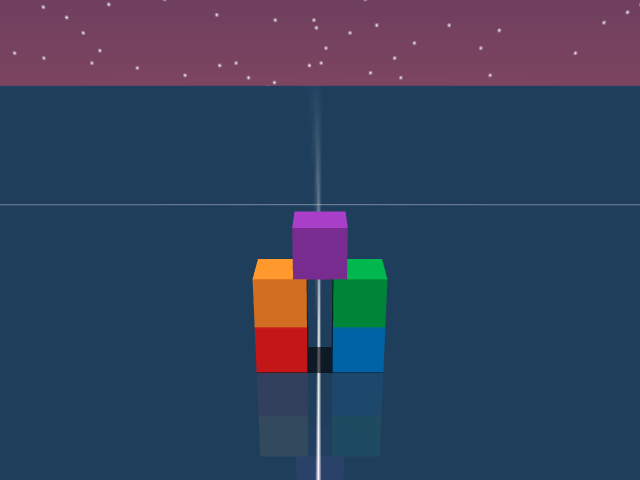} & Easy \\ \midrule
cube-5-task-4 & \includegraphics[width=4cm, margin=0pt 6pt 0pt 6pt, valign=c]{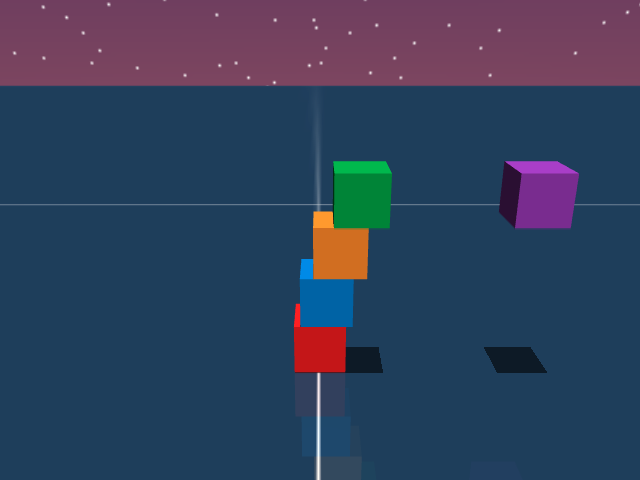} & Hard \\ \midrule
cube-5-task-5 & \includegraphics[width=4cm, margin=0pt 6pt 0pt 6pt, valign=c]{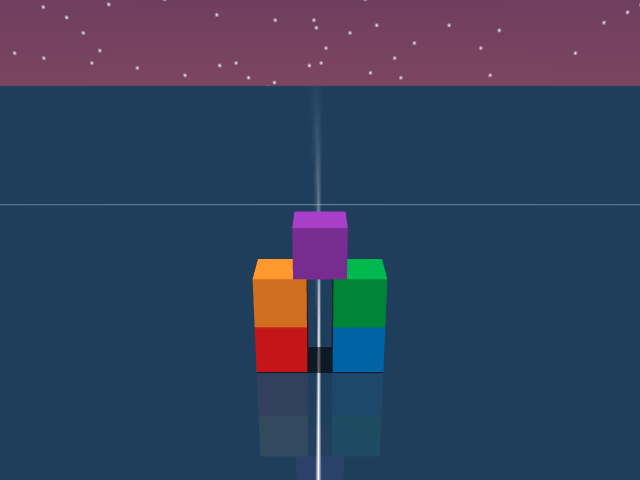} & Easy \\ \midrule

cube-6-task-1 & \includegraphics[width=4cm, margin=0pt 6pt 0pt 6pt, valign=c]{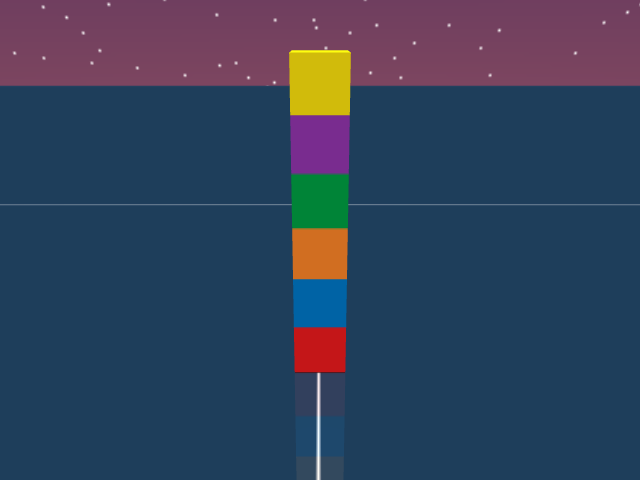} & Easy \\ \midrule
cube-6-task-2 & \includegraphics[width=4cm, margin=0pt 6pt 0pt 6pt, valign=c]{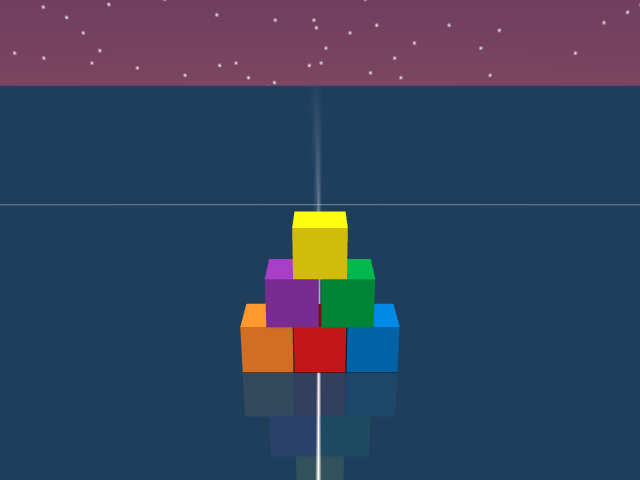} & Easy \\ \midrule
cube-6-task-3 & \includegraphics[width=4cm, margin=0pt 6pt 0pt 6pt, valign=c]{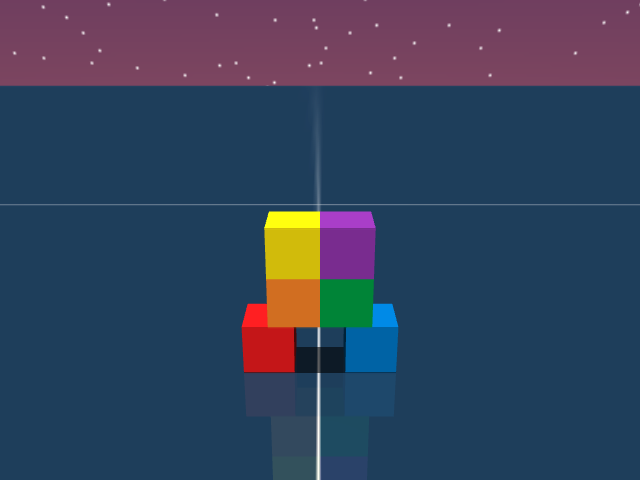} & Hard \\ \midrule
cube-6-task-4 & \includegraphics[width=4cm, margin=0pt 6pt 0pt 6pt, valign=c]{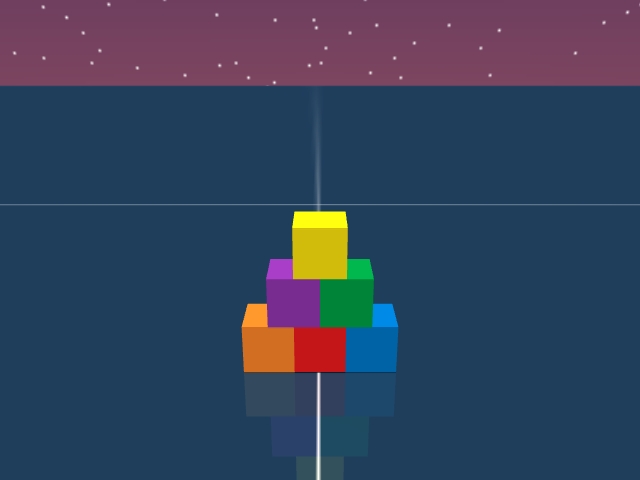} & Easy \\ \midrule
cube-6-task-5 & \includegraphics[width=4cm, margin=0pt 6pt 0pt 6pt, valign=c]{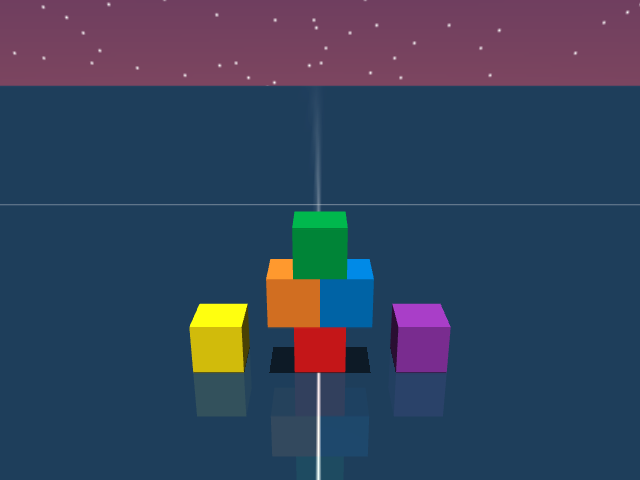} & Hard \\ \midrule

cube-7-task-1 & \includegraphics[width=4cm, margin=0pt 6pt 0pt 6pt, valign=c]{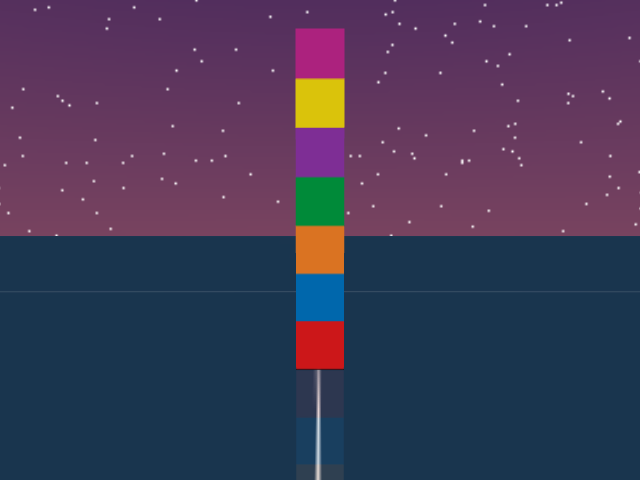} & Easy \\ \midrule
cube-7-task-2 & \includegraphics[width=4cm, margin=0pt 6pt 0pt 6pt, valign=c]{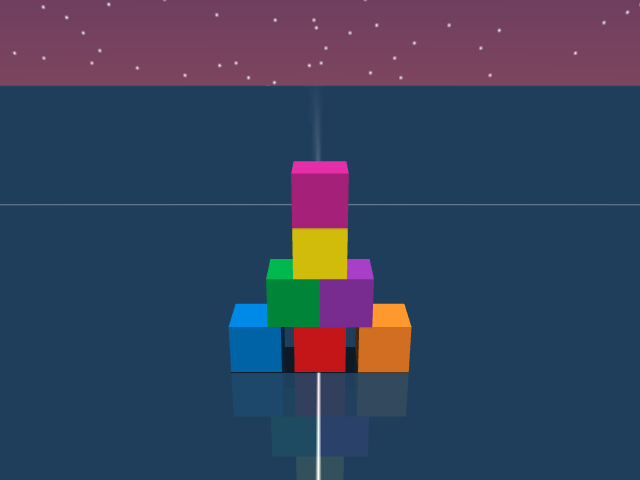} & Easy \\ \midrule
cube-7-task-3 & \includegraphics[width=4cm, margin=0pt 6pt 0pt 6pt, valign=c]{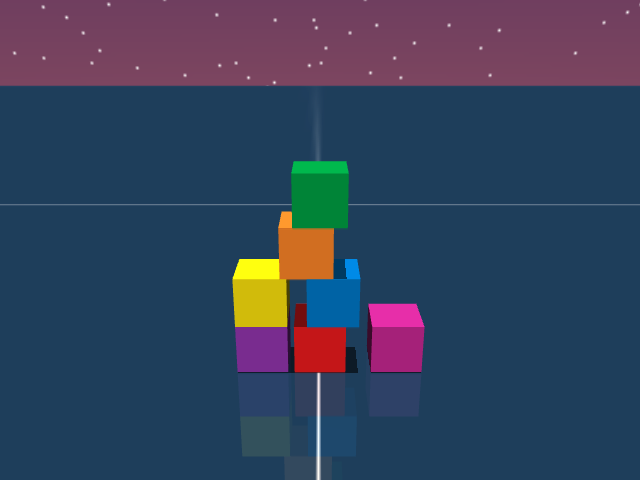} & Easy \\ \midrule
cube-7-task-4 & \includegraphics[width=4cm, margin=0pt 6pt 0pt 6pt, valign=c]{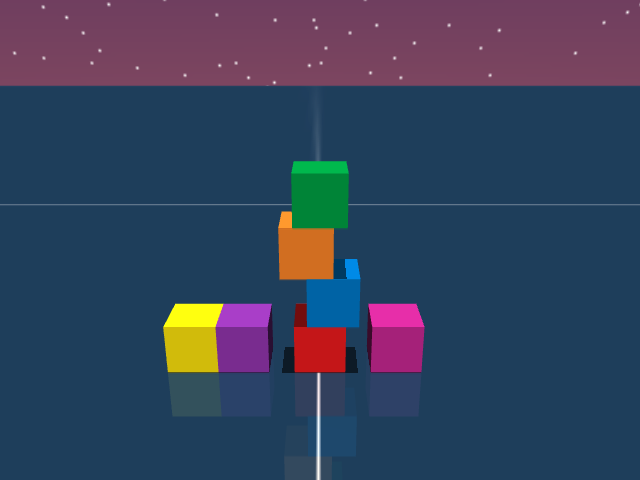} & Hard \\ \midrule
cube-7-task-5 & \includegraphics[width=4cm, margin=0pt 6pt 0pt 6pt, valign=c]{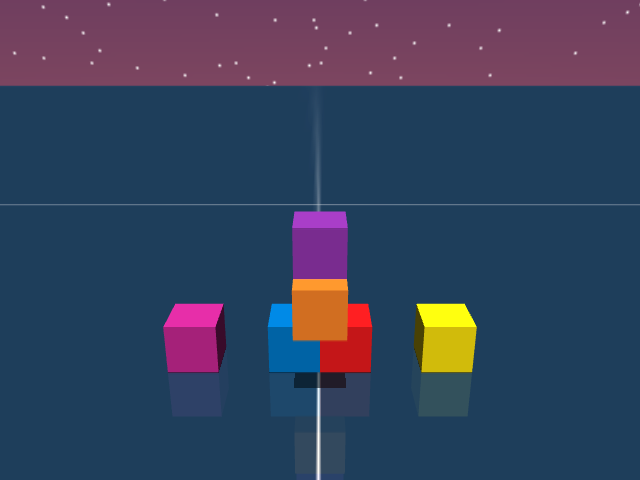} & Hard \\ \midrule
cube-7-task-6 & \includegraphics[width=4cm, margin=0pt 6pt 0pt 6pt, valign=c]{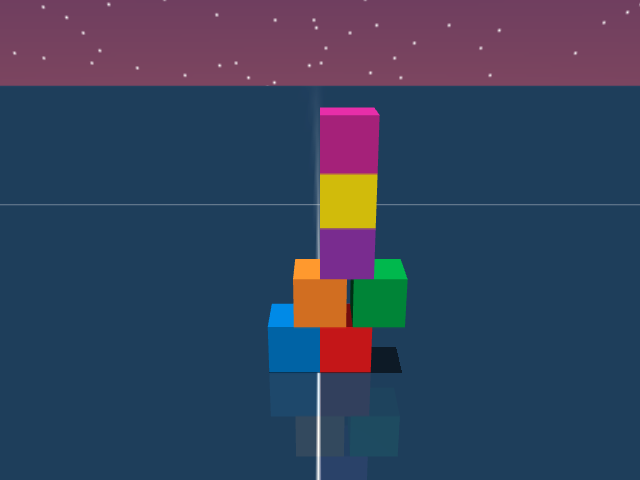} & Hard \\ \midrule
cube-7-task-7 & \includegraphics[width=4cm, margin=0pt 6pt 0pt 6pt, valign=c]{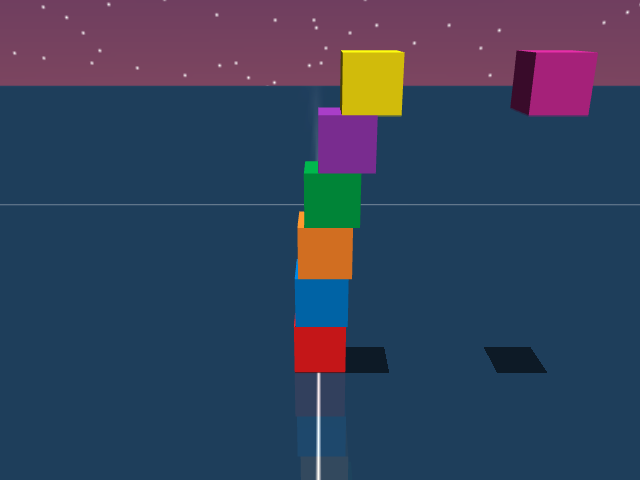} & Hard \\ \midrule

cube-8-task-1 & \includegraphics[width=4cm, margin=0pt 6pt 0pt 6pt, valign=c]{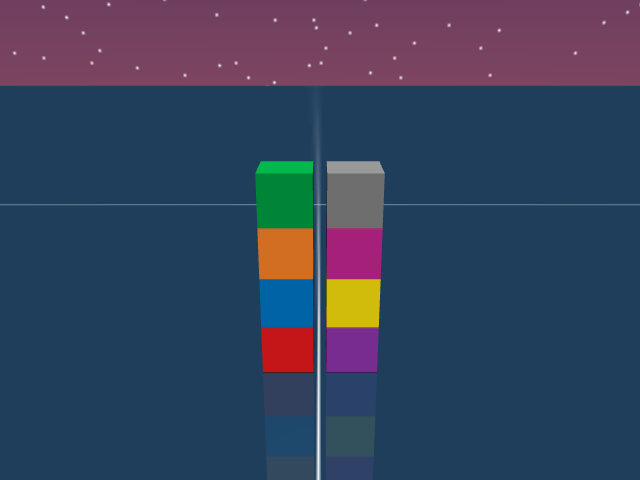} & Easy \\ \midrule
cube-8-task-2 & \includegraphics[width=4cm, margin=0pt 6pt 0pt 6pt, valign=c]{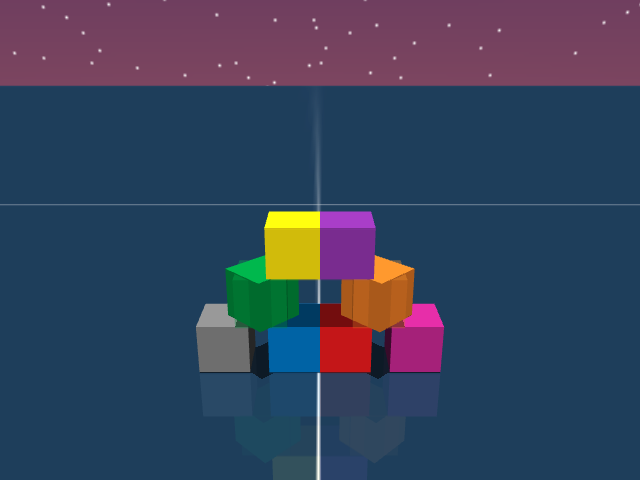} & Easy \\ \midrule
cube-8-task-3 & \includegraphics[width=4cm, margin=0pt 6pt 0pt 6pt, valign=c]{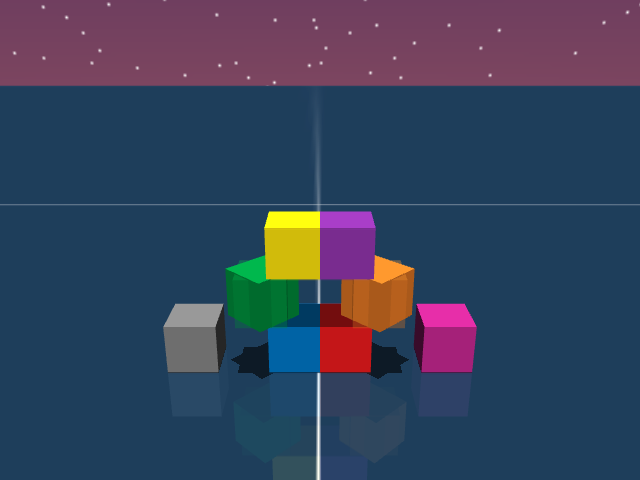} & Hard \\ \midrule
cube-8-task-4 & \includegraphics[width=4cm, margin=0pt 6pt 0pt 6pt, valign=c]{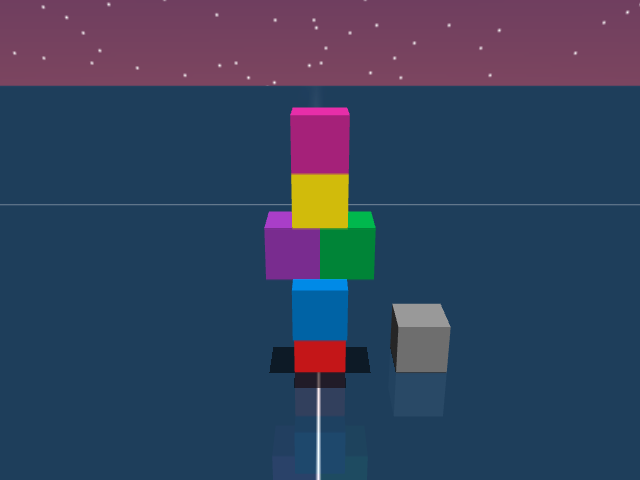} & Hard \\ \midrule
cube-8-task-5 & \includegraphics[width=4cm, margin=0pt 6pt 0pt 6pt, valign=c]{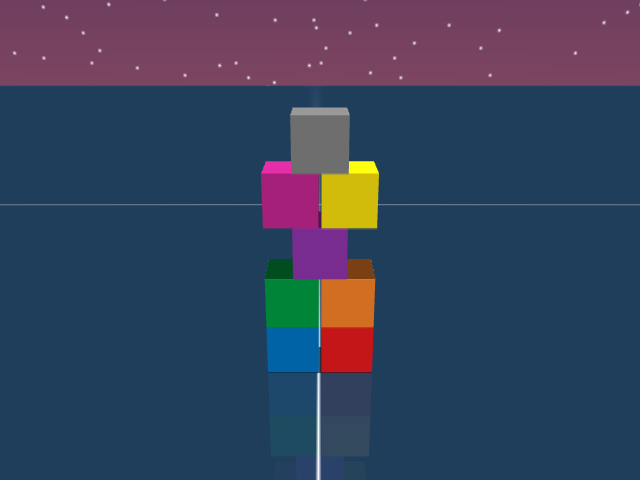} & Hard \\ \midrule

cube-9-task-1 & \includegraphics[width=4cm, margin=0pt 6pt 0pt 6pt, valign=c]{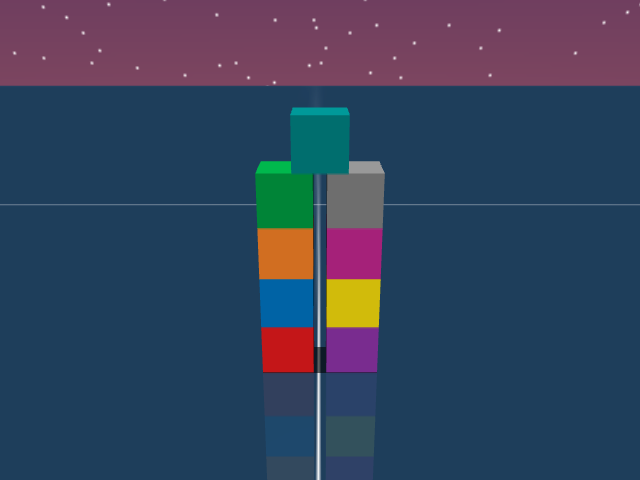} & Easy \\ \midrule
cube-9-task-2 & \includegraphics[width=4cm, margin=0pt 6pt 0pt 6pt, valign=c]{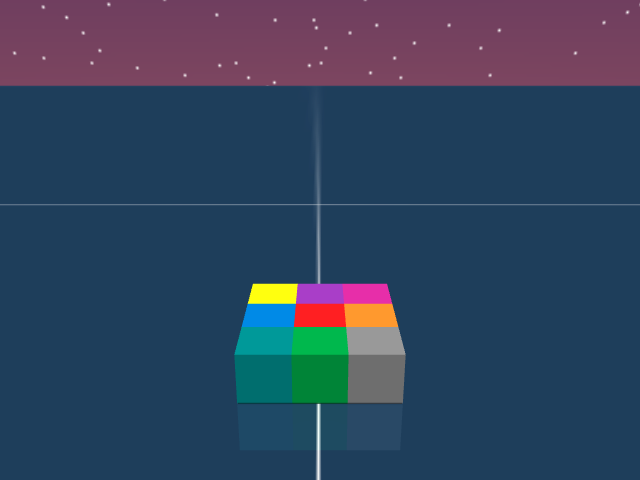} & Hard \\ \midrule
cube-9-task-3 & \includegraphics[width=4cm, margin=0pt 6pt 0pt 6pt, valign=c]{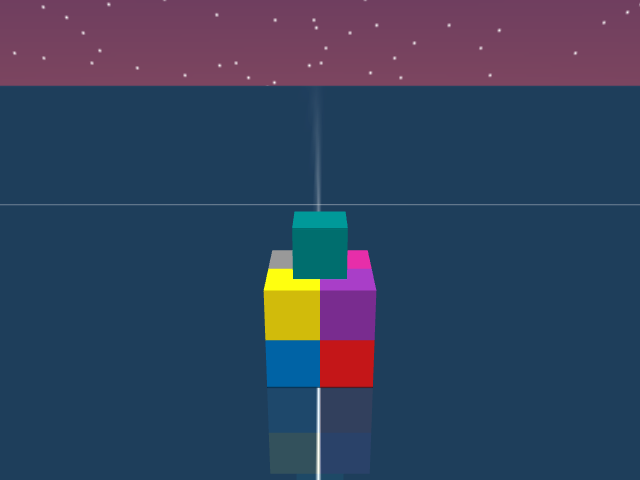} & Hard \\ \midrule
cube-9-task-4 & \includegraphics[width=4cm, margin=0pt 6pt 0pt 6pt, valign=c]{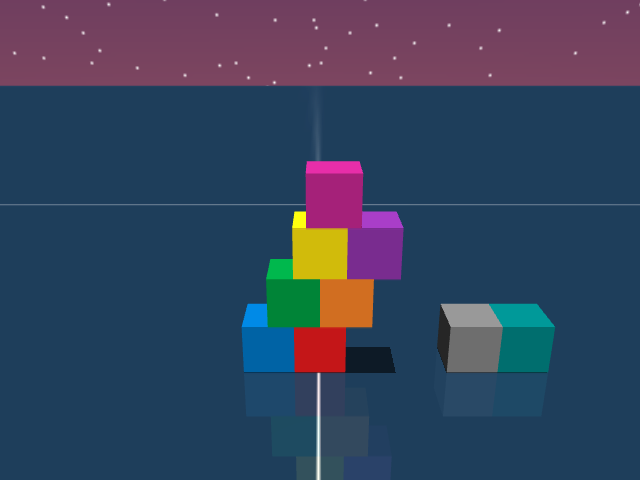} & Hard \\ \midrule

cube-10-task-1 & \includegraphics[width=4cm, margin=0pt 6pt 0pt 6pt, valign=c]{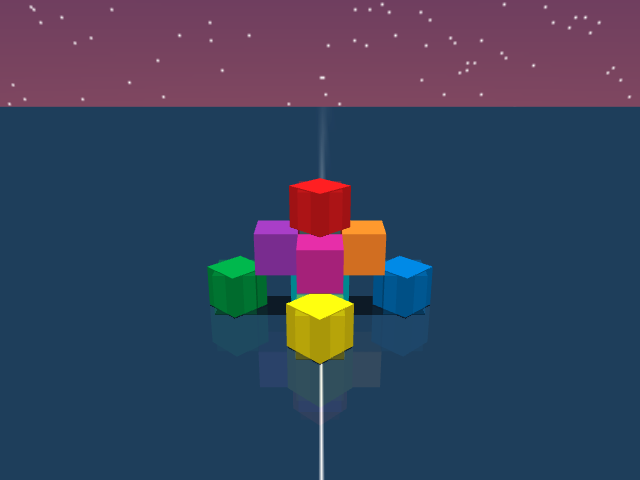} & Hard \\ \midrule

cube-15-task-1 & \includegraphics[width=4cm, margin=0pt 6pt 0pt 6pt, valign=c]{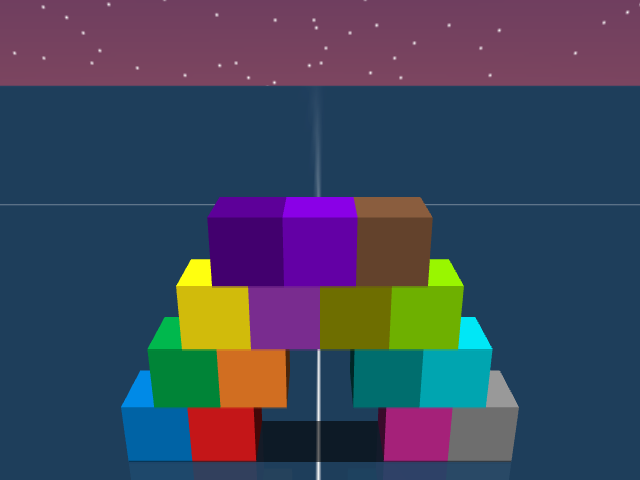} & Hard \\ \midrule
cube-15-task-2 & \includegraphics[width=4cm, margin=0pt 6pt 0pt 6pt, valign=c]{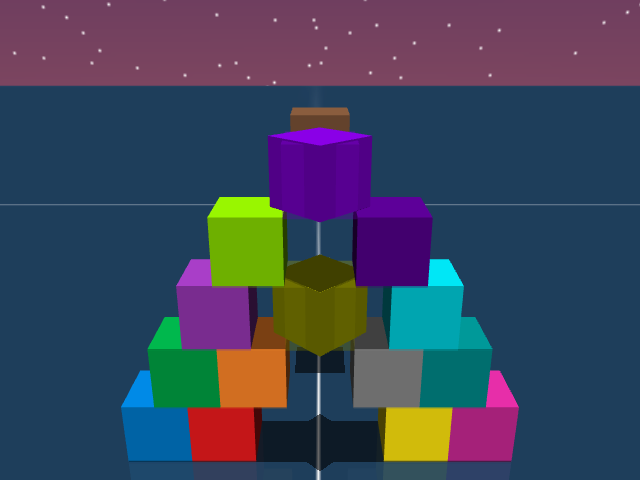} & Hard \\ \midrule

cube-20-task-1 & \includegraphics[width=4cm, margin=0pt 6pt 0pt 6pt, valign=c]{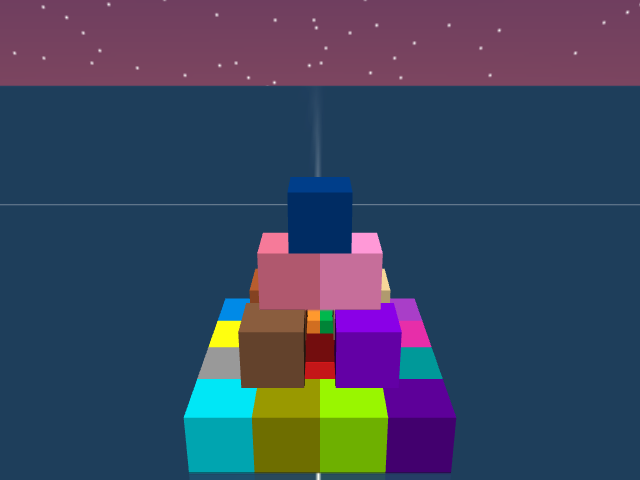} & Hard \\ \midrule

cube-50-task-1 & \includegraphics[width=4cm, margin=0pt 6pt 0pt 6pt, valign=c]{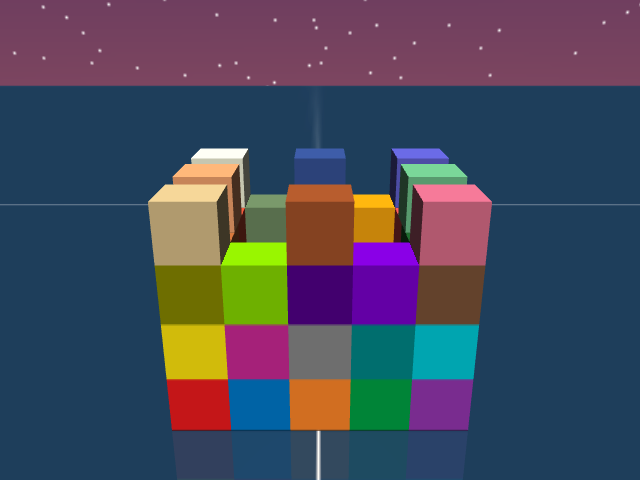} & Hard \\ \midrule
cube-50-task-2 & \includegraphics[width=4cm, margin=0pt 6pt 0pt 6pt, valign=c]{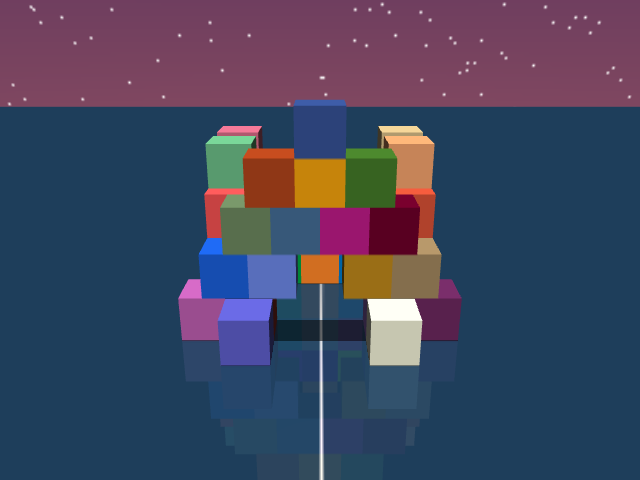} & Hard \\ 

\end{longtable}

\newpage
\section*{NeurIPS Paper Checklist}

\begin{enumerate}

\item {\bf Claims}
    \item[] Question: Do the main claims made in the abstract and introduction accurately reflect the paper's contributions and scope?
    \item[] Answer: \answerYes{}
    \item[] Justification: As noted in the abstract, the paper open sources the code for the benchmark containing a simulator and a task suite of over 50 tasks. We also show the results of both the frontier model based agents and tabula rasa RL agents.

\item {\bf Limitations}
    \item[] Question: Does the paper discuss the limitations of the work performed by the authors?
    \item[] Answer: \answerYes{}
    \item[] Justification: Limitations are discussed in~\cref{sec:limitations}

\item {\bf Theory assumptions and proofs}
    \item[] Question: For each theoretical result, does the paper provide the full set of assumptions and a complete (and correct) proof?
    \item[] Answer: \answerNA{}
    \item[] Justification: There are no theoretical results in this paper.

    \item {\bf Experimental result reproducibility}
    \item[] Question: Does the paper fully disclose all the information needed to reproduce the main experimental results of the paper to the extent that it affects the main claims and/or conclusions of the paper (regardless of whether the code and data are provided or not)?
    \item[] Answer:  \answerYes{}
    \item[] Justification: Code for the benchmark, all baselines, and experiments is included \href{https://github.com/neurips-submission-2026-abedstwsd99232/BuilderBench}{here}. 

\item {\bf Open access to data and code}
    \item[] Question: Does the paper provide open access to the data and code, with sufficient instructions to faithfully reproduce the main experimental results, as described in supplemental material?
    \item[] Answer: \answerYes{} 
    \item[] Justification: Code for the benchmark, all baselines, and experiments is included \href{https://github.com/neurips-submission-2026-abedstwsd99232/BuilderBench}{here}. The code contains a detailed README file that contains instructions to faithfully reproduce all the experimental results.

\item {\bf Experimental setting/details}
    \item[] Question: Does the paper specify all the training and test details (e.g., data splits, hyperparameters, how they were chosen, type of optimizer) necessary to understand the results?
    \item[] Answer: \answerYes{} 
    \item[] Justification: The exact details of all the experiments are provided in the code. All algorithms have their own clean single file implementation where the exact hyperparameters are stated.

\item {\bf Experiment statistical significance}
    \item[] Question: Does the paper report error bars suitably and correctly defined or other appropriate information about the statistical significance of the experiments?
    \item[] Answer: Yes 
    \item[] Justification: All figures have complete statistical information of the experiments.

\item {\bf Experiments compute resources}
    \item[] Question: For each experiment, does the paper provide sufficient information on the computer resources (type of compute workers, memory, time of execution) needed to reproduce the experiments?
    \item[] Answer: \answerYes{}{} 
    \item[] Justification: \Cref{app:compute-usage} provides details about the compute resources used in our experiments.
    
\item {\bf Code of ethics}
    \item[] Question: Does the research conducted in the paper conform, in every respect, with the NeurIPS Code of Ethics \url{https://neurips.cc/public/EthicsGuidelines}?
    \item[] Answer: \answerYes{} 
    \item[] Justification: The research conducted in the paper conforms in every respect, with the NeurIPS Code of Ethics

\item {\bf Broader impacts}
    \item[] Question: Does the paper discuss both potential positive societal impacts and negative societal impacts of the work performed?
    \item[] Answer: \answerNA{} 
    \item[] Justification: This work provides a benchmark for exploration and learning via interaction. The experiments use available language models and existing RL algorithms. Hence, this work does not pose a direct negative societal impact.
    
\item {\bf Safeguards}
    \item[] Question: Does the paper describe safeguards that have been put in place for responsible release of data or models that have a high risk for misuse (e.g., pre-trained language models, image generators, or scraped datasets)?
    \item[] Answer: \answerNA{} 
    \item[] Justification: To the best of our knowledge, the paper does not pose a high risk for misuse.

\item {\bf Licenses for existing assets}
    \item[] Question: Are the creators or original owners of assets (e.g., code, data, models), used in the paper, properly credited and are the license and terms of use explicitly mentioned and properly respected?
    \item[] Answer: \answerYes{} 
    \item[] Justification: We have cited original owners of assets or provided URLs wherever necessary.

\item {\bf New assets}
    \item[] Question: Are new assets introduced in the paper well documented and is the documentation provided alongside the assets?
    \item[] Answer: \answerYes{} 
    \item[] Justification: The code for the benchmark and the experiments is well documented and contains a comprehensive README file.

\item {\bf Crowdsourcing and research with human subjects}
    \item[] Question: For crowdsourcing experiments and research with human subjects, does the paper include the full text of instructions given to participants and screenshots, if applicable, as well as details about compensation (if any)? 
    \item[] Answer: \answerNA{} 
    \item[] Justification: the paper does not involve crowdsourcing nor research with human subjects.

\item {\bf Institutional review board (IRB) approvals or equivalent for research with human subjects}
    \item[] Question: Does the paper describe potential risks incurred by study participants, whether such risks were disclosed to the subjects, and whether Institutional Review Board (IRB) approvals (or an equivalent approval/review based on the requirements of your country or institution) were obtained?
    \item[] Answer: \answerNA{} 
    \item[] Justification: The paper does not involve crowdsourcing nor research with human subjects

\item {\bf Declaration of LLM usage}
    \item[] Question: Does the paper describe the usage of LLMs if it is an important, original, or non-standard component of the core methods in this research? Note that if the LLM is used only for writing, editing, or formatting purposes and does \emph{not} impact the core methodology, scientific rigor, or originality of the research, declaration is not required.
    \item[] Answer: \answerYes{} 
    \item[] Justification: The experiment section~(\Cref{sec:experiments}) describes how different LLMs are evaluated in our experiments.

\end{enumerate}

\end{document}